\documentclass[10pt,twocolumn,letterpaper]{article}

\usepackage{cvpr}              

\usepackage[table]{xcolor} 

\newcommand\correspondingauthor{\thanks{Corresponding author. This work is supported by NExT++ Research Center and National Research Foundation and Singapore under its AI Singapore Programme (AISG Award No: AISG2-RP-2021-022).}}
\definecolor{cvprblue}{rgb}{0.21,0.49,0.74}
\definecolor{mygray}{gray}{.9}

\usepackage[pagebackref,breaklinks,colorlinks,citecolor=cvprblue]{hyperref}
\usepackage{soul}
\usepackage{epigraph}
\usepackage{amsmath}

\usepackage{multirow, tabularx}
\usepackage{makecell}
\usepackage{booktabs} 
\usepackage{arydshln} 
\usepackage{array} 
\usepackage{xspace} 
\usepackage{diagbox}
\usepackage{bbding} 
\usepackage{subfloat}

\newcommand{\mysubsubsec}[1]{\noindent$\bullet$~\textbf{#1}}
\newcommand{\pub}[1]{{\color{gray}{\tiny{[{#1}]}}}}
\newcommand{\qua}[1]{{\color{gray}{\footnotesize{#1}}}} 

\makeatletter
\newcommand{\thickhline}{
	\noalign {\ifnum 0=`}\fi \hrule height 1pt
	\futurelet \reserved@a \@xhline
}
\makeatother

\def\paperID{5535} 
\def\confName{CVPR}
\def\confYear{2024}

\title{Discriminative Probing and Tuning for Text-to-Image Generation}

\author{Leigang Qu$^1$, Wenjie Wang$^1$\correspondingauthor, Yongqi Li$^2$, Hanwang Zhang$^{3,4}$, Liqiang Nie$^5$, Tat-Seng Chua$^1$\\
\normalsize$^1$National University of Singapore, 
$^2$Hong Kong Polytechnic University,
\normalsize$^3$Nanyang Technological University, \\
\normalsize$^4$Skywork AI, 
\normalsize$^5$Harbin Institute of Technology (Shenzhen)\\
{\tt\small leigangqu@gmail.com, wenjiewang96@gmail.com, liyongqi0@gmail.com} \\
{\tt\small hanwangzhang@ntu.edu.sg, nieliqiang@gmail.com, dcscts@nus.edu.sg}
}

\begin{document}
\maketitle
\begin{abstract}
Despite advancements in text-to-image generation (T2I), prior methods often face text-image misalignment problems such as relation confusion in generated images. 
Existing solutions involve cross-attention manipulation for better compositional understanding or integrating large language models for improved layout planning. 
However, the inherent alignment capabilities of T2I models are still inadequate. 
By reviewing the link between generative and discriminative modeling, we posit that T2I models' discriminative abilities may reflect their text-image alignment proficiency during generation. 
In this light, we advocate bolstering the discriminative abilities of T2I models to achieve more precise text-to-image alignment for generation. 
We present a discriminative adapter built on T2I models to probe their discriminative abilities on two representative tasks and leverage discriminative fine-tuning to improve their text-image alignment. 
As a bonus of the discriminative adapter, a self-correction mechanism can leverage discriminative gradients to better align generated images to text prompts during inference. 
Comprehensive evaluations across three benchmark datasets, including both in-distribution and out-of-distribution scenarios, demonstrate our method's superior generation performance. 
Meanwhile, it achieves state-of-the-art discriminative performance on the two discriminative tasks compared to other generative models. 
The code is available at \url{https://dpt-t2i.github.io/}. 
\end{abstract}

\section{Introduction}
\label{sec:intro}

\begin{figure}[t]
\centering
\includegraphics[width=0.45\textwidth]{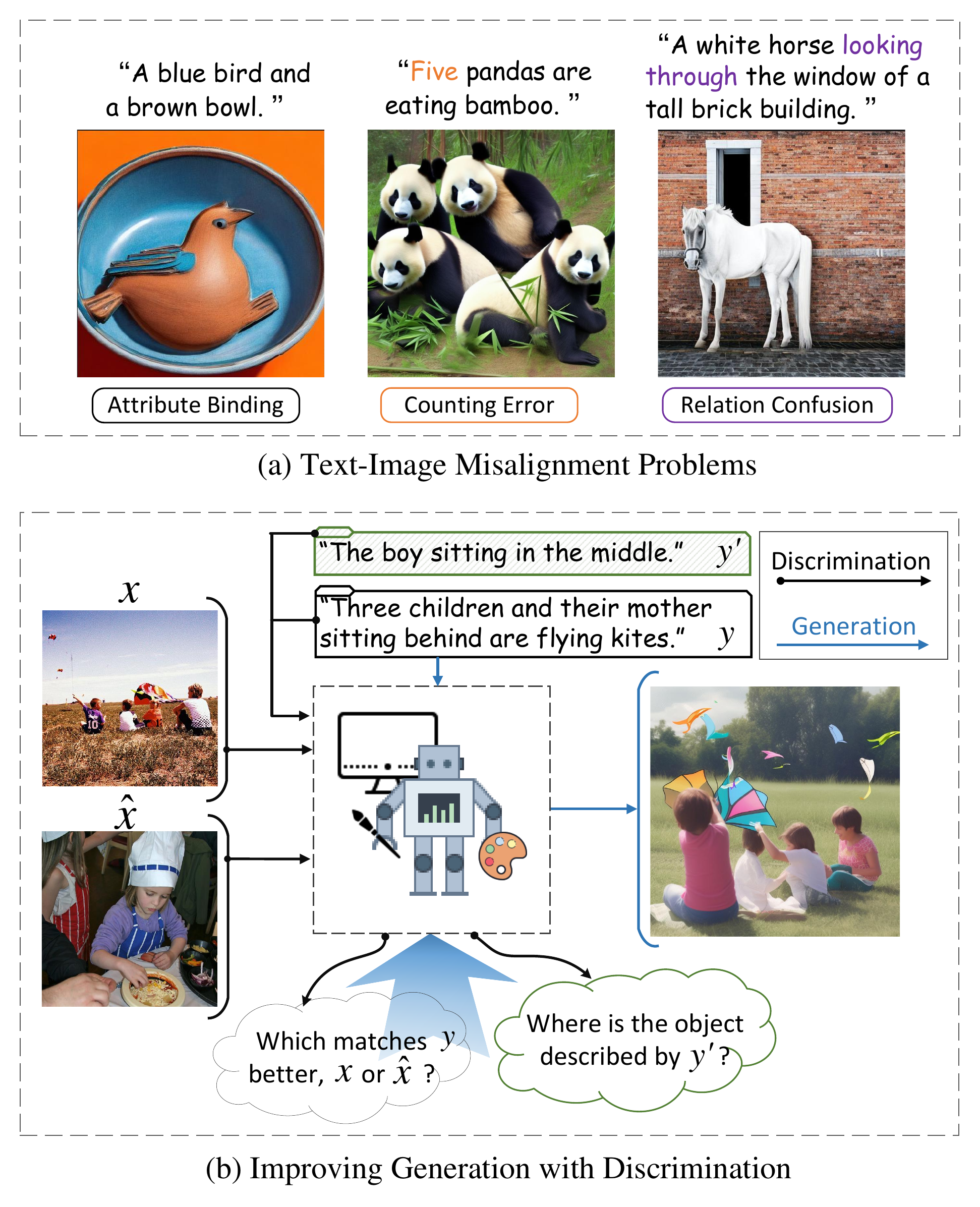}
\vspace{-2ex}
\caption{Illustration of the (a) text-image misalignment problem and (b) our motivation by enhancing discriminative abilities of T2I models to promote generative abilities. We list three wrong generation results generated by SD-v2.1~\cite{rombach2022high} with regard to attribute binding, counting error, and relation confusion in (a). }
\label{fig:gen_dist}
\vspace{-3ex}
\end{figure}

Text-to-image generation (T2I) aims to synthesize high-quality and semantically-relevant images to a given free-form text prompt. 
In recent years, the rapid development of diffusion models~\cite{sohl2015deep, ho2020denoising} has ignited the research enthusiasm for content generation, leading to a significant leap in T2I~\cite{ramesh2022hierarchical, saharia2022photorealistic, rombach2022high}. 
However, due to the weak compositional reasoning capabilities, current T2I models still suffer from the \textbf{Text-Image Misalignment} problem~\cite{lee2023aligning}, such as attribute binding~\cite{feng2022training}, counting error~\cite{qu2023layoutllm}, and relation confusion~\cite{qu2023layoutllm} (see Fig.~\ref{fig:gen_dist}), especially in complicated multi-object generation scenes. 

Two lines of work have made remarkable progress in improving text-image alignment for T2I models. The first line proposes to intervene in cross-modal attention activations guided by linguistic structures~\cite{feng2022training} or test time optimization~\cite{chefer2023attend}. 
However, they heavily rely on the inductive bias for manipulating attention structures, often necessitating expertise in vision-language interaction. This expertise is not easily acquired and lacks flexibility.
In contrast, another research line~\cite{qu2023layoutllm, feng2023layoutgpt} borrows LLM's linguistic comprehension and compositional abilities for layout planning, and then incorporates layout-to-image models (\eg, GLIGEN~\cite{li2023gligen}) for controllable generation. 
Although these methods mitigate misalignment issues like counting error, they heavily rely on intermediate states, \eg, bounding boxes, for layout representation. The intermediate states may not adequately capture fine-grained visual attributes, and can also accumulate errors in this two-stage paradigm. Furthermore, the intrinsic compositional reasoning abilities of T2I models are still inadequate.

To tackle these issues, we aim to promote text-image alignment by directly catalyzing the intrinsic compositional reasoning of T2I models, without depending on the inductive bias for attention manipulation or intermediate states. 
Richard Feynman famously stated, ``What I cannot create, I do not understand,'' underscoring the significance of understanding in the process of creation. This motivates us to consider enhancing the understanding abilities of T2I models to facilitate their text-to-image generation. As illustrated in Fig.~\ref{fig:gen_dist}, T2I models are more likely to generate an image with correct semantics if they can distinguish the alignment difference between the text prompt and the two images with minor semantic variations. 

In light of this, we propose to examine the understanding abilities of T2I models by two discriminative tasks. 
First, we probe the discriminative \textit{global matching} ability\footnote{Here we inspect the understanding ability of models with discriminative tasks by considering the taxonomy of discriminative and generative learning in Machine Learning.} of T2I models on Image-text Matching (ITM)~\cite{frome2013devise, qu2021dynamic}, a representative task to evaluate fundamental text-image alignment. 
The second discriminative task inspects the \textit{local grounding} ability of T2I models. One representative task is Referring Expression Comprehension (REC)~\cite{yu2016modeling}, which examines the fine-grained expression-object alignment within an image. 
Based on the two tasks, we aim to 1) probe the discriminative abilities of T2I models, especially the compositional semantic alignment, and 2) further improve their discriminative abilities for better text-to-image generation.

Toward this end, we propose a \underline{D}iscriminative \underline{P}robing and \underline{T}uning (\textbf{DPT}) paradigm to examine and improve text-image alignment of T2I models in a two-stage process. 
1) To probe the discriminative abilities, DPT incorporates a Discriminative Adapter to do the ITM and REC tasks based on the semantic representations~\cite{kwon2022diffusion} of T2I models. 
For example, DPT may take the feature maps from U-Net of diffusion models~\cite{rombach2022high} as semantic representations. 
And 2) in the second stage, DPT further improves the text-image alignment by means of parameter-efficient fine-tuning, \eg, LoRA~\cite{hu2021lora}. In addition to the adapter, DPT fine-tunes the foundation T2I models to strengthen its intrinsic compositional reasoning abilities for both discriminative and generative tasks. 
As an extension, we present a self-correction mechanism to guide T2I models for better alignment by gradient-based guidance signals from the discriminative adapter. 
We conduct extensive experiments on three alignment-oriented text-to-image generation benchmarks
and four ITM and REC benchmarks 
under in-distribution and out-of-distribution settings, validating the effectiveness of DPT in enhancing both generative and discriminative abilities of T2I models. 
The main contributions of this work are threefold. 
\begin{itemize}
    \item We retrospect the relations between generative and discriminative modeling, and propose a simple yet effective paradigm called DPT to probe and improve the basic discriminative abilities of T2I models for better text-to-image generation. 
    \item We present a discriminative adapter to achieve efficient probing and tuning in DPT. Besides, we extend T2I models with a self-correction mechanism guided by the discriminative adapter for alignment-oriented generation. 
    \item We conduct extensive experiments on three text-to-image generation datasets and four discriminative datasets, 
    significantly enhancing the generative and discriminative abilities of representative T2I models. 
\end{itemize}

\section{Related Work}
\label{sec:related_work}

\noindent \textbf{$\bullet$ Text-to-Image Generation}.
Over the past decades, great efforts on Variational Autoencoders~\cite{yan2016attribute2image}, Generative Adversarial Networks~\cite{zhang2017stackgan, xu2018attngan}, and Auto-regression Models~\cite{ramesh2021zero, ding2021cogview, yu2022scaling} have been dedicated to generating high-quality images with text conditions. Recently, there has been a flurry of interest in Diffusion Probabilistic Models (DMs)~\cite{sohl2015deep, ho2020denoising} due to their stability and scalability. To further improve the generation quality, large-scale models such as DALL$\cdot$E 2~\cite{ramesh2022hierarchical}, Imagen~\cite{saharia2022photorealistic}, and GLIDE~\cite{nichol2021glide}, emerged to synthesize photorealistic images. 
This work mainly focuses on diffusion models and especially takes the open-sourced Stable Diffusion (SD)~\cite{rombach2022high} as the base model.

\vspace{3pt}
\noindent \textbf{$\bullet$ Improving Text-Image Alignment}. 
Despite the thrilling success, current T2I models still suffer from Text-Image Misalignment issues~\cite{gokhale2022benchmarking, conwell2022testing, bakr2023hrs}, especially in complex scenes requiring compositional reasoning~\cite{liu2022compositional}. 
Several pioneering efforts were made to introduce guidance to intervene in internal features of SD to stimulate the high-alignment generation. 
For example, StructureDiffusion~\cite{feng2022training} parses prompts into tree structures and incorporates them with cross-attention representations to promote compositional generation.  
Attend-and-Excite~\cite{chefer2023attend} manipulates cross-attention units to attend to all textual subject tokens and enhance the activations in attention maps. 
Despite the notable momentum, they are limited to tackling problems including missing objects and incorrect attributes, and ignore relation enhancement. 
Another thread of work, \eg, LayoutLLM-T2I~\cite{qu2023layoutllm} and LayoutGPT~\cite{feng2023layoutgpt}, resorts to two-stage coarse-to-fine frameworks~\cite{gupta2021layouttransformer, fan2023frido, razavi2019generating}, in which they first induce explicit intermediate bounding box-based layout, and then synthesize images. However, such an intermediate layout may not be sufficient to represent complex scenes and they almost abandon the intrinsic reasoning abilities of pre-trained T2I models. 
In this work, we propose a discriminative tuning paradigm by stimulating discriminative abilities of pre-trained T2I models for high-alignment generation. 

\vspace{3pt}
\noindent \textbf{$\bullet$ Generative and Discriminative Modeling}.
The thrilling progress of LLMs enables generative models to complete discriminative tasks, which motivates researchers to exploit understanding abilities~\cite{lin2023multi} with foundation visual generative models in Image Classification~\cite{li2023your, clark2023text, chen2023robust, yang2023diffusion}, Segmentation~\cite{burgert2022peekaboo, xu2023open, zhao2023unleashing}, and Image-Text Matching~\cite{krojer2023diffusion}. Besides, DreamLLM~\cite{dong2023dreamllm} unifies generation and discrimination in a multimodal auto-regressive framework and reveals the potential synergy. On the contrary, a recent work~\cite{west2023generative} discusses the generative AI paradox and showed LLMs may not indeed understand what they have generated. To the best of our knowledge, we are the first to study discriminative tuning to promote alignment in T2I.

\begin{figure*}[t]
\centering
	\includegraphics[width=0.98\textwidth]{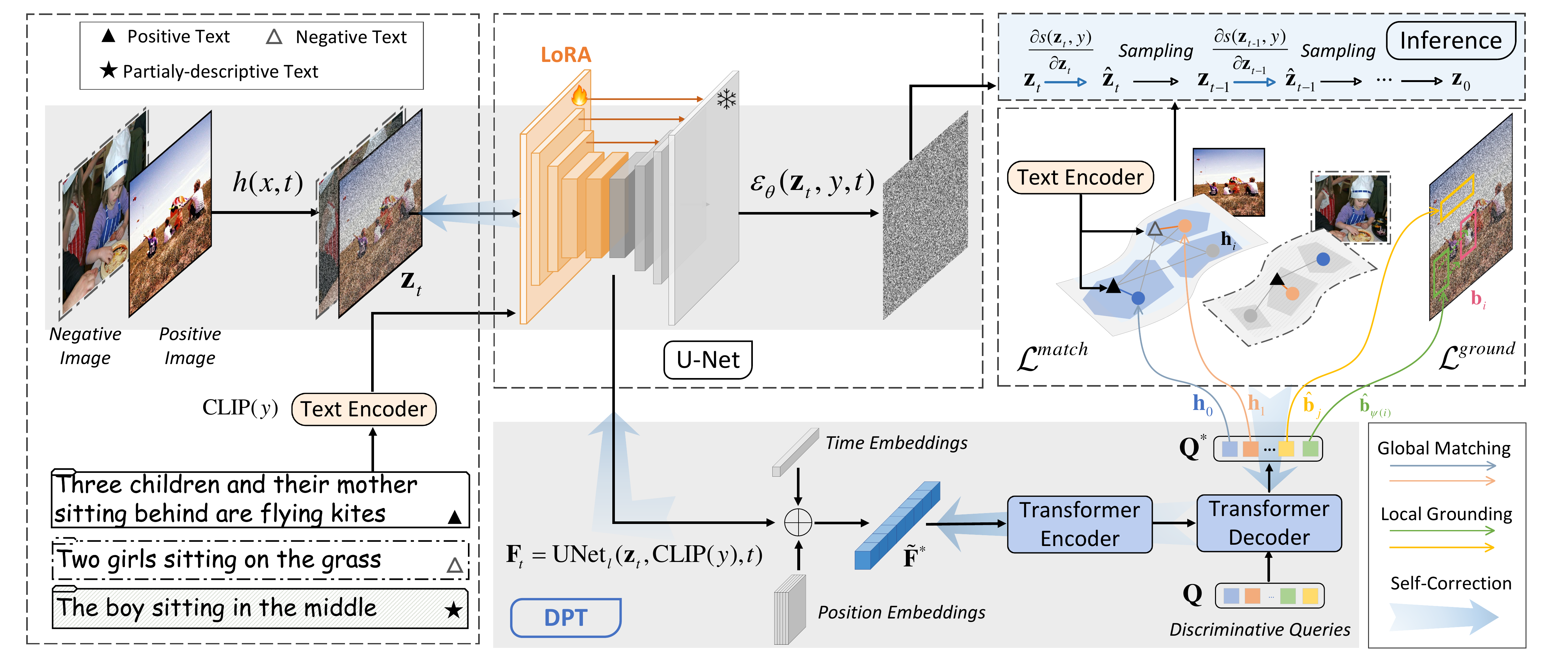}
	\vspace{-1ex}
	\caption{Schematic illustration of the proposed discriminative probing and tuning (DPT) framework. We first extract semantic representations from the frozen SD and then propose a discriminative adapter to conduct discriminative probing to investigate the global matching and local grounding abilities of SD. Afterward, we perform parameter-efficient discriminative tuning by introducing LoRA parameters. During inference, we present the self-correction mechanism to guide the denoising-based text-to-image generation. } 
	\label{fig:pipeline}
	\vspace{-2ex}
\end{figure*}

\section{Method}

In this section, we introduce the DPT paradigm to probe and enhance the discriminative abilities of foundation T2I models. 
As shown in Fig.~\ref{fig:pipeline}, DPT consists of two stages, \ie, Discrimination Probing and Discrimination Tuning, as well as a self-correction mechanism in Sec.~\ref{sec:self_corr}. 

\subsection{Stage 1 -- Discriminative Probing}\label{sec:dis_prob}
In the first stage, we aim to develop a probing method to explore
\textit{``How powerful are discriminative abilities of recent T2I models?''}. To this end, we first select representative T2I models and semantic representations, and then consider adapting the T2I models to do discriminative tasks. 

\vspace{3pt}
\mysubsubsec{Stable Diffusion for Discriminative Probing.}
Considering SD is open-sourced and one of the most powerful and popular T2I models, we select its different versions (see Sec.~\ref{sec:exp_per}) as representative models to probe the discriminative abilities. 
To make generative diffusion models semantically focused and efficient, SD~\cite{rombach2022high} performs denoising in a latent low-dimensional space. It includes VAE~\cite{kingma2013auto}, Text Encoder of CLIP~\cite{radford2021learning}, and U-Net~\cite{ronneberger2015u}. 
The U-Net serves as a neural backbone for denoising score matching in the latent space, composed of three parts, \ie, down blocks, mid blocks, and up blocks. 
During training, given a positive image-text pair $(x, y)$, SD first encodes image $x$ with the VAE encoder and adds noise $\epsilon \sim \mathcal{N}(0, 1)$ to obtain the latent $\mathbf{z}_t = h(x, t)$ at timestep $t$. Thereafter, SD employs U-Net to predict the added noise and optimizes the model parameters by minimizing the L2 loss between the ground-truth noise and the predicted one.

\mysubsubsec{Semantic Representations.} 
It is non-trivial to leverage T2I models such as SD to do discriminative tasks. Fortunately,
recent work~\cite{kwon2022diffusion} demonstrates that diffusion models have a meaningful semantic latent space although they were originally designed for denoising~\cite{ho2020denoising} or score estimation~\cite{song2020score}. 
Besides, a series of pioneering work~\cite{li2023your, clark2023text, burgert2022peekaboo, xu2023open} shows the validity and even superiority of representations extracted from U-Net of SD to be qualified to discriminative tasks. 
Inspired by these studies, we consider utilizing semantic representations from the U-Net of SD to do discriminative tasks via a discriminative adapter. 

\vspace{3pt}
\mysubsubsec{Discriminative Adapter}.
We propose a lightweight discriminative adapter, which relies on the semantic representations of SD to handle discriminative tasks. 
Inspired by DETR~\cite{carion2020end}, we implement the discriminative adapter with the Transformer~\cite{vaswani2017attention} structure, including a Transformer encoder and a Transformer decoder. Besides, we adopt a fixed number of randomly initialized and learnable queries to adapt the framework to specific discriminative tasks. 

Concretely, given a noisy latent $\mathbf{z}_t$
at a sampled timestep $t$ and a prompt $y$, 
we first feed them into U-Net and extract a 2D feature map $\mathbf{F}_t \in \mathbb{R}^{h \times w \times d}$ from one of the intermediate blocks\footnote{We select the medium block by default, and also delve into the influence of different blocks in Sec.~\ref{sec:exp_in_depth}.}, where $h$, $w$, and $d$ denote the height, width, and dimension, respectively. Formally, we extract $\mathbf{F}_t$ via
\begin{equation}
\small
    \mathbf{F}_t = {\rm UNet}_l (\mathbf{z}_t, {\rm CLIP} (y), t), 
\end{equation}
where ${\rm UNet}_l$ refers to the operation of extracting the feature maps in the $l$-th block of U-Net.
Afterward, we combine $\mathbf{F}_t$ with learnable position embeddings~\cite{dosovitskiy2020image} and timestep embeddings~\cite{rombach2022high} of $t$ via additive fusion, and then flatten it into the semantic representation $\tilde{\mathbf{F}}_t \in \mathbb{R}^{hw \times d}$. 
For simplicity, we will omit the subscript $t$ in the following. 

To probe the discriminative abilities, we feed $\tilde{\mathbf{F}}$ into the Transformer encoder ${\rm Enc(\cdot)}$, and then perform interaction between the encoder output and some learnable queries $\mathbf{Q} = \{\mathbf{q}_1, ..., \mathbf{q}_N\}$ with $q_i \in \mathbb{R}^d$ in the Transformer decoder ${\rm Dec(\cdot, \cdot)}$. The whole process is formulated as 
\begin{equation}
\small
\mathbf{Q}^* = f(\tilde{\mathbf{F}};\mathcal{W}_a, \mathbf{Q}) = {\rm Dec}({\rm Enc} (\tilde{\mathbf{F}}), \mathbf{Q})
\end{equation}
where $f(\cdot)$ abstracts to the discriminative adapter with parameters $\mathcal{W}_a$ and $\mathbf{Q}$.  $\mathcal{W}_a$ includes the parameters in ${\rm Enc}$ and ${\rm Dec}$. 
The queries $\mathbf{Q}$ serve as a bridge between visual representations and downstream discriminative tasks, which attends the encoded semantic representation $\tilde{\mathbf{F}}_t$ via cross-attention~\cite{vaswani2017attention} of the decoder for downstream tasks.  
Thanks to multiple queries in $\mathbf{Q}$, the query representations $\mathbf{Q}^*$ 
capture multiple aspects of the semantic representation $\tilde{\mathbf{F}}$. 
Thereafter, $\mathbf{Q}^*$ can be used to do various downstream tasks, possibly with a classier or regressor.  

In the following, we will introduce two probing tasks, \ie, ITM and REC, and train the discriminative adapter on them to investigate the global matching and local grounding abilities of T2I models, respectively.

\vspace{3pt}
\mysubsubsec{Global Matching}.
From the view of discriminative modeling, a model with strong text-image alignment should be able to identify subtle alignment differences between various images and a text prompt. 
In light of this, We utilize the task of Image-Text Matching~\cite{frome2013devise} to probe the discriminative global matching ability. This task is defined to achieve bidirectional matching or retrieval, including text-to-image (${T \rightarrow I}$) and image-to-text (${I \rightarrow T}$). 

To achieve this, we first collect the first $M (M < N)$ query representations $\{\mathbf{q}^*_1, ..., \mathbf{q}^*_M\}$ from $\mathbf{Q}^*$, and then project each of them into a matching space with the same dimension as CLIP and obtain $\mathbf{h}_i = g(\mathbf{q}^*_i;\mathcal{W}_m)$. 
Intuitively, different query representations may capture different aspects to understand the same image. 
Inspired by this, 
we calculate the cross-modal semantic similarities between $x$ and $y$ by comparing the CLIP textual embedding of $y$ and the most matched projected query representations via $s(y, \mathbf{z}) = \max_{i \in \{1, ..., M\}} \cos({\rm CLIP}(y), \mathbf{h}_i)$. 
Based on pairwise similarities, we optimize the discriminative adapter $f(\cdot; \mathcal{W}_a, \mathbf{Q})$ and the projection layer $g(\cdot; \mathcal{W}_m)$ using contrastive learning loss $\mathcal{L}^{match} = \mathcal{L}^{T \rightarrow I} + \mathcal{L}^{I \rightarrow T}$. 
The first term optimizes the model to distinguish the correct image matched with a given text from all samples in a batch, \ie,  
\begin{equation}\label{eqn:loss_t2i}
\small
\mathcal{L}^{T \rightarrow I} = -\log \frac{\exp (s(\mathbf{z}, y) / \tau)}{\sum_{j=1}^B  \exp s((\mathbf{z}_j, y) / \tau)}, 
\end{equation}
where $B$ denotes the min-batch size, and $\tau$ is a learnable temperature factor. Similarly, the opposite direction from image to text is computed by
\begin{equation}\label{eqn:loss_i2t}
\small
\mathcal{L}^{I \rightarrow T} = -\log \frac{\exp (s(\mathbf{z}, y) / \tau)}{\sum_{j=1}^B \exp s((\mathbf{z}, y_j) / \tau)}.
\end{equation}

With $\mathcal{L}^{match}$ as the optimization objective, the discriminative adapter and the projection layers are enforced to discover discriminative information from the semantic representations for matching, implying the global matching ability of a T2I model. 

\vspace{3pt}
\mysubsubsec{Local Grounding}.
Local grounding
requires a model to recognize the referred object from others in an image given a partially descriptive text. 
We adapt SD to the REC~\cite{yu2016modeling} task to evaluate its discriminative local grounding ability.

Formally, given a textual expression $y'$ referring to a specific object with index $i$ in an image $x$, REC aims to predict the coordinate and the size, \ie, the bounding box $\mathbf{b}_i$, of the ground-truth object. 
To achieve it, we share the same discriminative adapter and employ the other $(N - M)$
learnable queries as object prior queries and obtain the corresponding query representations from the transformer decoder as $\{\mathbf{q}^*_j\}_{j \in\{M+1, ..., N\}}$.
We then project each $\mathbf{q}^*_j$ into three spaces separately by three different project layers $g(\cdot)$: 1) the grounding space to 
get the probability of predicting 
the correct object, \ie, $p_j = g(\mathbf{q}^*_j;\mathcal{W}_p) \in \mathbb{R}^1$; 2) the box space to estimate the bounding box parameters, \ie, $\hat{\mathbf{b}}_j = g(\mathbf{q}^*_j; \mathcal{W}_b) \in \mathbb{R}^4$; 
and 3) the semantic space to bridge the semantic gap between queries and the text, \ie, $\mathbf{o}_j = g(\mathbf{q}^*_j; \mathcal{W}_s) \in \mathbb{R}^d$. 

After projection, we perform maximum matching to discover the most matched query with index $\psi(i)$. The cost used for matching includes using grounding probability, L1, and GIoU~\cite{rezatofighi2019generalized} losses between the prediction and the ground-truth box as costs. It is formulated as 
\begin{equation}
\small
\begin{split}\label{eqn:max_matching}
    \psi(i) = \mathop{\arg\min}\limits_{j \in \{M+1, ..., N\}} - p_j +  {\rm L1}(\hat{\mathbf{b}}_j, \mathbf{b}_i) + {\rm GIoU} (\hat{\mathbf{b}}_j, \mathbf{b}_i)
\end{split}
\end{equation}
Besides, we adopt a text-to-object contrastive loss to further drive the model to distinguish the positive object from others at the semantic level:
\begin{equation}
\small
\mathcal{L}^{T \rightarrow O} = -\log \frac{\exp (\cos(\mathbf{o}_{\psi(i)}, {\rm CLIP}(y')) / \tau)}{\sum_{j=1}^{K_x}  \exp (\cos(\mathbf{o}_j, {\rm CLIP}(y')) / \tau)}, 
\end{equation}
We combine all the losses and obtain the grounding loss as 
\begin{equation}\label{eqn:grounding_loss}
\small
\begin{aligned}
\mathcal{L}^{ground} = &-\lambda_0 p_{\psi(i)} + \lambda_1 {\rm L1}(\hat{\mathbf{b}}_{\psi(i)}, \mathbf{b}_i) \\
&+ \lambda_2 {\rm GIoU} (\hat{\mathbf{b}}_{\psi(i)}, \mathbf{b}_i) 
+ \lambda_3 \mathcal{L}^{T \rightarrow O},  
\end{aligned}
\end{equation}
where $\{\lambda_k\}_{k \in \{0, 1, 2, 3\}}$ serve as trade-off factors.

Finally, we optimize the parameters of the whole model, including $\mathbf{Q}$ and $\{\mathcal{W}_i\}, i \in \{a, p, b, s\}$, with the following loss function on two tasks:
\begin{equation}
\small
    \mathcal{L} = \mathbb{E}_{x, \epsilon \sim \mathcal{N}(0,1), t} (\mathcal{L}_t^{match} + \mathcal{L}_t^{ground})
\end{equation}
The probing process includes training and inference on the two discriminative tasks. 
During training, we freeze all parameters of SD, and adopt its semantic representations for matching and grounding by optimizing the discriminative adapter and several projection layers. 
During inference, we obtain the testing performance on the two discriminative tasks, which reflects the discriminative abilities of SD.


\subsection{Stage 2 -- Discriminative Tuning}\label{sec:dis_tune}
In the second stage, we propose to improve the generative abilities, especially text-image alignment, by optimizing T2I models in a discriminative tuning manner. 
Most prior work~\cite{burgert2022peekaboo, xu2023open} only views SD as a fixed feature extractor for segmentation tasks due to its fine-grained semantic representation power but overlooks the potential back-feeding of discrimination to generation. 
Besides, though a recent study~\cite{krojer2023diffusion, xiang2023denoising} fine-tunes the SD model using discriminative objectives, it only pays attention to specific downstream tasks (\eg, ITM) and ignores the effect of tuning on generation. The advancement of discrimination may sacrifice the original generative power. 
In this stage, we mainly focus on enhancing generation, but also investigate the superior limit of discrimination under the premise of priority generation. 
It may shed new light on giving full play to the versatility of visual generative foundation models. 
In this vein, we strive to explain \textit{``How can we enhance text-image alignment for T2I models by discriminative tuning?''}

In the previous stage, we freeze SD and probe how informative intermediate activations are in global matching and local grounding. Here, we conduct parameter-efficient fine-tuning using LoRA~\cite{hu2021lora} by injecting trainable layers over cross-attention layers and freezing the parameters of the pre-trained SD. We use the same discriminative objective functions as stage 1 to tune the LoRA,  discriminative adapter, and task-specific projection layers. 
Due to the participation of LoRA, we can flexibly manipulate the intermediate activation of T2I models.

\subsection{Self-Correction}\label{sec:self_corr}
Equipping the T2I model with the discriminative adapter enables the whole model to execute discriminative tasks. 
As a bonus of using the discriminative adapter, 
we propose a self-correction mechanism to guide high-alignment generation during inference. 
Formally, we update the latent $\mathbf{z}_t$ 
aiming to enhance 
the semantic similarity between $\mathbf{z}_t$ and the prompt $y$ through gradients:  
\begin{equation}\label{eqa:self_corr}
\small
    \hat{\mathbf{z}}_t = \mathbf{z}_t + \eta \frac{\partial s(\mathbf{z}_t, y)}{\partial \mathbf{z}_t}, 
\end{equation}
where the guidance factor $\eta$ 
control the guidance strength. 
$\frac{\partial s(\mathbf{z}_t, y)}{\partial \mathbf{z}_t}$ represents the gradients from the discriminative adapter to the latent $\mathbf{z}_t$. 
Afterward, we predict the noise by feeding $\hat{\mathbf{z}}_t$ into U-Net and then obtain $\mathbf{z}_{t-1}$ for generation.

\section{Experiments}
We conduct extensive experiments to evaluate the generative and discriminative performance of DPT, justify its effectiveness, and conduct an in-depth analysis.

\begin{table*}[t]
	\centering
	\setlength{\abovecaptionskip}{0.15cm}
	\caption{Performance comparison for \textit{text-to-image generation} on COCO-NSS1K, CC-500, and ABC-6K. ID, OOD, and MD refer to in-distribution, out-of-distribution, and mixed-distribution settings, respectively.  According to the version of Stable Diffusion, we split methods into two groups, top and down for v1.4 and v2.1, respectively. SC denotes self-correction. }
	\label{tab:perf_comp_gen}
	{\hspace{-1ex}
             \setlength{\tabcolsep}{1mm}{
		\resizebox{0.97\textwidth}{!}
		{
			\setlength\tabcolsep{5pt}
			\renewcommand\arraystretch{1.1}
			\begin{tabular}{p{3cm}<{\raggedleft}|ccccc|ccccc|cccc}
				\hline\thickhline
				\multicolumn{1}{c|}{\multirow{2}{*}{Method}} & \multicolumn{5}{c|}{ COCO-NSS1K (ID)}  & \multicolumn{5}{c|}{ CC-500 (OOD) } & \multicolumn{4}{c}{ ABC-6K (MD)} \\ 
				\cline{2-15}
				 & CLIP & BLIP-M  & BLIP-C & \color{gray}{IS} & \color{gray}{FID} & CLIP & BLIP-M & BLIP-C  & GLIP & \color{gray}{IS} & CLIP & BLIP-M & BLIP-C & \color{gray}{IS} \\
				\hline\hline
				\multicolumn{1}{l|}{{Stable Diffusion-v1.4}~\pub{CVPR22}~\cite{rombach2022high}}    & 33.27  & 67.96  & 39.48  & \qua{31.32}  & \qua{54.77}  & 34.82  & 70.95  & 40.36  & 31.17 & \qua{14.28}  & 35.33  & 72.03  & 40.82  & \qua{34.47}  \\
                \multicolumn{1}{l|}{{LayoutLLM-T2I}~\pub{ACMMM23}~\cite{qu2023layoutllm}}    & 32.42  & 67.42  & 39.46  & \qua{25.57}  & \qua{59.26}  & -     & -     & -     & -     & \qua{-}     & -     & -     & -     & \qua{-} \\
                \multicolumn{1}{l|}{{StructureDiffusion}~\pub{ICLR23}~\cite{feng2022training}} & -     & -     & -     & \qua{-}     & \qua{-}     & 33.71  & 66.71  & 39.54  & 31.39 & \qua{14.14} &      34.95 & 69.55 & 40.69 & \qua{34.97} \\
                \multicolumn{1}{l|}{{HN-DiffusionITM}~\pub{NeurIPS23}~\cite{krojer2023diffusion}}    & 33.26  & 70.06  & \textbf{40.14}  & \qua{31.53}  & \qua{53.26}  & 34.15  & 68.77  & 40.30  & 31.54  & \qua{13.99}  & 35.02  & 72.28  & 41.12  & \qua{34.83}  \\
                \rowcolor{mygray}
                \multicolumn{1}{l|} {DPT (Ours)} & \textbf{33.85}  & \textbf{71.84}  & 40.11  & \qua{31.65}  & \qua{54.96}  & \textbf{35.97}  & \textbf{76.74}  & \textbf{41.15}  & \textbf{37.07} & \qua{13.56}    & \textbf{35.88}  & \textbf{75.88}  & \textbf{41.26}  & \qua{34.46}  \\
				\cdashline{1-15}[1pt/1pt]
                \multicolumn{1}{l|}{{Stable Diffusion-v2.1}~\pub{CVPR22}~\cite{rombach2022high}}    & 34.96  & 73.32  & 40.22  & \qua{30.40}  & \qua{55.35}  & 39.24  & 85.45  & 43.36  & 52.09  & \qua{11.53}  & 37.53  & 81.98  & 41.77  & \qua{33.31}  \\
                \multicolumn{1}{l|}{{Attend-and-
                 Excite}~\pub{TOG23}~\cite{chefer2023attend}}      &  34.95  & 74.68  & 40.32  & \qua{30.27}  & \qua{55.16}  & 39.43  & 90.03  & 44.08  & \textbf{53.29}  & \qua{11.82}  & 37.59 & 82.64  & 41.83  & \qua{32.94}  \\
                \multicolumn{1}{l|}{{HN-DiffusionITM}~\pub{NeurIPS23}~\cite{krojer2023diffusion}}    & 35.14  & 75.64  & 40.77  & \qua{30.34}  & \qua{52.73}  & 38.81  & 85.76  & 43.22  & 48.95  & \qua{12.11}  & 37.58  & 82.33  & 42.07  & \qua{34.14}  \\
                \rowcolor{mygray}
				\multicolumn{1}{l|} {DPT (Ours)} & \textbf{35.83}  & \underline{78.58}  & \underline{41.14}  & \qua{30.83}  & \qua{55.55}  & \underline{40.23}  & \underline{90.72}  & \underline{44.55}  & \textbf{53.29}  & \qua{11.59}  & \underline{38.39}  & \textbf{86.19}  & \textbf{42.36}  & \qua{32.97}  \\
                \rowcolor{mygray}
                \multicolumn{1}{l|} {DPT + SC (Ours)} & \underline{35.75}  & \textbf{79.15}  & \textbf{41.14}  & \qua{30.50}  & \qua{54.89}  & \textbf{40.25}  & \textbf{91.33}  & \textbf{44.69}  & \textbf{53.29}  & \qua{11.89}  & \textbf{38.41}  & \underline{85.63}  & \underline{42.34}  & \qua{33.56} \\
				\hline
			\end{tabular}
		}
            }
	}
\end{table*}

\begin{table}[h]
	\centering
	\setlength{\abovecaptionskip}{0.15cm}
	\caption{Performance comparison for \textit{text-to-image generation} on TIFA~\cite{hu2023tifa} and T2I-CompBench~\cite{huang2023t2i}. According to the version of Stable Diffusion, we split methods into two groups, top and down for v1.4 and v2.1, respectively. SC denotes self-correction. }
    \label{tab:tifa_t2i_compb}
	{\hspace{-1.3ex}
		\resizebox{0.47\textwidth}{!}
		{
			\setlength\tabcolsep{3.7pt}
			\renewcommand\arraystretch{1.3}
			\begin{tabular}{l|p{1cm}<{\centering}|cccccc}
                \hline\thickhline
				&  \multicolumn{1}{c|}{\multirow{2}{*}{TIFA}}  & \multicolumn{6}{c}{T2I-CompBench} \\
                \cline{3-8}
                & & Color & Shape & Text. & Sp. & Non-Sp. & Comp. \\
				\hline
                SD-v1.4~\cite{rombach2022high} & 79.15 & 36.82  & 35.94  & 42.16  & 10.64  & 30.45  & 28.18  \\
                HN-DiffusionITM~\cite{krojer2023diffusion} & 79.02 & 36.71  & 35.48  & 39.84  & 11.22  & \textbf{30.91}  & 28.05  \\
                VPGen~\cite{cho2023visual} & 77.33 & 32.12  & 32.36  & 35.85  & 19.08  & 30.07  & 24.39  \\
                LayoutGPT~\cite{feng2023layoutgpt} & 79.31  & 33.86  & 36.35  & 44.07  & \textbf{35.06}  & 30.30  & 26.36  \\
                \rowcolor{mygray}
                DPT (Ours) & 82.04 & 48.84  & 38.93  & 50.10  & 14.63  & 30.83  & 30.05  \\
                \rowcolor{mygray}
                DPT + SC (Ours) & \textbf{82.40} & \textbf{51.51}  & \textbf{39.61}  & \textbf{49.38}  & 15.45  & 30.84  & \textbf{30.29}  \\
                \hline
                SD-v2.1~\cite{rombach2022high} & 81.35 & 48.21  & 40.49  & 46.83  & 16.94  & 30.63  & 29.96  \\
                Attend-and-Excite~\cite{chefer2023attend} & 81.98  & 53.72  & 43.41  & 48.53  & 16.30  & 30.64  & 30.38  \\
                HN-DiffusionITM~\cite{krojer2023diffusion} & 82.02 & 46.45  & 40.09  & 49.35  & 15.01  & \textbf{30.99}  & 30.35  \\
                \rowcolor{mygray}
                DPT (Ours) & \underline{84.49} & \underline{60.59}  & \underline{48.18}  & \textbf{58.24}  & \underline{20.78}  & \underline{30.95}  & \underline{32.44}  \\
                \rowcolor{mygray} 
                DPT + SC (Ours) & \textbf{84.63} & \textbf{62.59}  & \textbf{48.44}  & \underline{57.60}  & \textbf{21.04}  & 30.76  & \textbf{32.52}  \\
                \hline
			\end{tabular}
		}
	}
\vspace{-0.3cm}
\end{table}

\subsection{Experimental Settings}
\noindent \mysubsubsec{Benchmarks}. During training, we adopt the training set of MSCOCO~\cite{lin2014microsoft} for ITM and three commonly used datasets~\cite{yu2016modeling}, \ie, RefCOCO, RefCOCO+, and RefCOCOg for REC. 
To evaluate the text-image alignment, we utilize five benchmarks: COCO-NSS1K~\cite{qu2023layoutllm}, CC-500~\cite{feng2022training}, ABC-6K~\cite{feng2022training}, TIFA~\cite{hu2023tifa}, and T2I-CompBench~\cite{huang2023t2i}. 
According to the distribution differences of textual prompts between the training set and the test sets, we adopt three settings, \ie, In-Distribution (ID) and Out-of-Distribution (OOD)~\cite{sun2022counterfactual} on COCO-NSS1K and CC-500, respectively, and Mixed Distribution (MD) on ABC-6K, TIFA, and T2I-CompBench. 
More details can be found in Appendix \ref{sec:app_benchmarks}.

\mysubsubsec{Evaluation Metrics}. Following the existing baselines~\cite{feng2022training, chefer2023attend, qu2023layoutllm}, we adopt CLIP score~\cite{hessel2021clipscore} and BLIP score\footnote{OpenCLIP (ViT-H-14)~\cite{cherti2023reproducible} and BLIP-2 (pretrain) are used to compute text-image similarities as CLIP and BLIP scores, respectively. We will adopt Image-Text Matching (ITM) and Image-Text Contrastive (ITC) as BLIP scores in the following. }~\cite{li2023blip} including BLIP-ITM and BLIP-ITC, and GLIP score~\cite{feng2022training} based on object detection to evaluate text-image alignment, and IS~\cite{salimans2016improved} and FID~\cite{heusel2017gans} as quality evaluation metrics. As for TIFA and T2I-CompBench, we follow the recommended VQA accuracy or specifically curated protocols.

\begin{table*}[t]
	\centering
	\setlength{\abovecaptionskip}{0.15cm}
	\caption{Performance comparison for \textit{image-text matching} and \textit{referring expression comprehension} to evaluate global matching and local grounding abilities, respectively. Datasets include MSCOCO-HN for ITM, and RefCOCO, RefCOCO+, and RefCOCOg for REC. All the methods are grouped into three parts, in which the upper, middle, and lower groups correspond to zero-shot discriminative, zero-shot generative, and fine-tuning generative methods, respectively. All the generative models are based on Stable Diffusion-v2.1. }
    \label{tab:perf_comp_dist}
	{\hspace{-1.3ex}
            \setlength{\tabcolsep}{0.5mm}{
		\resizebox{0.97\textwidth}{!}
		{
			\setlength\tabcolsep{10pt}
			\renewcommand\arraystretch{1.1}
			\begin{tabular}{p{3cm}<{\raggedleft}|ccc|ccc:c p{0.8cm}<{\centering}c:cc}
				\hline\thickhline
				\multicolumn{1}{c|}{\multirow{2}{*}{Method}} & \multicolumn{3}{c|}{ MSCOCO-HN} & \multicolumn{3}{c:}{ RefCOCO}  & \multicolumn{3}{c:}{ RefCOCO+} & \multicolumn{2}{c}{ RefCOCOg} \\ 
				\cline{2-12}
				& I-to-T & T-to-I & Overall & val & testA  & testB & val & testA & testB & val & test  \\
				\hline\hline
                \multicolumn{1}{l|} {Random Chance}       & 25.00  & 25.00  & 25.00  & 16.53  & 13.51  & 19.20  & 16.29  & 13.57  & 19.60  & 18.12  & 19.10  \\
				\multicolumn{1}{l|}{{CLIP (ViT-B-32)}~\pub{ICML21}~\cite{radford2021learning}} & 47.63  & 42.82  & 45.23  & 44.79  & 46.12  & 42.61  & 49.60  & 51.07  & 46.04  & 58.31  & 58.42  \\
                \multicolumn{1}{l|}{{OpenCLIP (ViT-B-32)}~\pub{CVPR23}~\cite{cherti2023reproducible}} & 49.07  & 47.45  & 48.26  & 43.22  & 43.15  & 44.65  & 48.21  & 48.60  & 50.64  & 60.32  & 60.84  \\
                \cdashline{1-12}[1pt/1pt]
                \multicolumn{1}{l|}{{Diffusion Classifier~$\S$}~\pub{ICCV23}~\cite{li2023your}} & 34.59  & 24.12  & 29.36  & 6.23  & 2.14  & 12.11  & 6.07  & 2.11  & 12.29  & 8.68  & 8.45  \\
                \multicolumn{1}{l|}{{DiffusionITM~$\S$}~\pub{NeurIPS23}~\cite{krojer2023diffusion}}      &  34.59  & 29.83  & 32.21  & 28.88  & 30.16  & 29.01  & 29.97  & 31.17  & 30.25  & 38.07  & 38.91  \\
                \multicolumn{1}{l|} {Local Dinoising}      & -     & -     & -     & 23.83  & 21.20  & 24.85  & 24.07  & 21.31  & 25.45  & 28.66  & 28.59  \\
                \cdashline{1-12}[1pt/1pt]
                \multicolumn{1}{l|}{{Diffusion Classifier~$\dagger\S$}~\pub{ICCV23}~\cite{li2023your}}  &37.72  & 24.03  & 30.88  & 6.11  & 2.10  & 10.91  & 6.04  & 2.13  & 11.48  & 8.05  & 7.54  \\
                \multicolumn{1}{l|}{{DiffusionITM~$\dagger\S$}~\pub{NeurIPS23}~\cite{krojer2023diffusion}}      &   37.72  & 29.88  & 33.80  & 34.09  & 32.70  & 35.29  & 35.86  & 35.42  & 38.23  & 49.67  & 49.05  \\
                \multicolumn{1}{l|}{{HN-DiffusionITM~$\dagger\S$}~\pub{NeurIPS23}~\cite{krojer2023diffusion}}      &  37.55  & 30.37  & 33.96  & 31.43  & 28.50  & 35.47  &   33.47    & 30.16  & 37.47  & 47.98  & 48.20  \\
                \multicolumn{1}{l|} {Local Dinoising~$\dagger$}      &  -     & -     & -     & 23.70  & 21.55  & 24.81  & 24.01  & 21.52  & 25.32  & 28.53  & 28.77  \\
                \rowcolor{mygray}
                \multicolumn{1}{l|} {DPT (Stage1, Ours)}       & 42.29  & 34.75  & 38.52  & 48.79  & 53.28  & 43.06  & 42.56  & 47.69  & 36.14  & 46.56  & 45.75  \\
                \rowcolor{mygray}
                \multicolumn{1}{l|} {DPT (Ours)}       & 42.07  & 34.97  & 38.52  & 52.73  & 57.84  & 46.73  & 45.34  & 50.12  & 38.41  & 48.61  & 47.45  \\
                \rowcolor{mygray}
                \multicolumn{1}{l|} {DPT* (Ours)}       &  \textbf{43.12}  & \textbf{35.25}  & \textbf{39.18}  & \textbf{63.45}  & \textbf{66.70}  & \textbf{57.90}  & \textbf{51.56}  & \textbf{56.81}  & \textbf{42.73}  & \textbf{54.96}  & \textbf{54.80}  \\
                \hline
                \multicolumn{10}{l}{\makecell[l]{$\dagger$: fine-tuning with the denoising objective;\\ 
                $\S$: cropping an image into blocks and then matching them with the referring text for REC; \\
                $*$: model selection with a priority discriminative task, \ie, ITM or REC}} 
			\end{tabular}
		}}
	}
         \vspace{-0.2cm}
\end{table*}

\subsection{Performance Comparison}\label{sec:exp_per}
\mysubsubsec{Text-to-Image Generation.}
As shown in Tab.~\ref{tab:perf_comp_gen} and Tab.~\ref{tab:tifa_t2i_compb}, we have the following observations and discussions: 1) Compared with the base foundation models, \ie, SD~\cite{rombach2022high}, the proposed DPT manages to improve the text-image alignment remarkably, which illustrates that enhancing discriminative abilities could benefit the generative semantic alignment for T2I models. 2) DPT achieves superior performance on CC-500 and ABC-6K under the OOD setting, showing its powerful generalization to other prompt distributions. It also reveals its capability to resist the risk of overfitting when tuning T2I models with discriminative tasks. 3) The consistent improvement on both SD-v1.4 and SD-v2.1 demonstrates that the proposed DPT may be parallel with the generative pre-training based on score matching, reflecting the possibility of activating the intrinsic reasoning abilities of T2I models using DPT. And 4) in all, the proposed method achieves the best generation performance consistently on text-image alignment across comprehensive benchmarks, distribution settings, and evaluation protocols. Besides, the improvement in alignment does not result in a loss of image quality per IS and FID. These results confirm the effectiveness of the proposed paradigm DPT.

\vspace{3pt}
\mysubsubsec{Discriminative Matching and Grounding}.
In Sec.~\ref{sec:dis_prob}, we incorporate a discriminative adapter on top of T2I models and probe and improve its understanding abilities based on ITM and REC. In an empirical sense, we carry out experiments by training the adapter in the first stage and introducing the LoRA for tuning in the second stage using ITM and REC data, and then evaluate the matching and grounding performance. We show experimental results of baselines including discriminative and generative models under the zero-shot and fine-tuning settings in Tab.~\ref{tab:perf_comp_dist}. See Appendix~\ref{sec:app_exp_settings} for more details on the implementation and settings. 
From this table, we observe that our method could outperform the existing state-of-the-art generative methods, such as Diffusion Classifier~\cite{li2023your} and DiffusionITM~\cite{krojer2023diffusion}, by large margins on ITM and REC tasks. Even it could achieve competitive performance in the first probing stage or when selected with a priority generation in the second stage. 
These results show that the generative representations extracted from the intermediate layers of U-Net convey meaningful semantics, verifying that T2I models have basic discriminative matching and grounding abilities. Besides, it also indicates that such abilities could be further improved by the discriminative tuning introduced in Sec.~\ref{sec:dis_tune}. 

\subsection{In-depth Analysis}\label{sec:exp_in_depth}
To verify the effectiveness of each component in DPT, including discriminative tuning on Global Matching (GM) and Local Grounding (LG) in the 2nd stage, and the Self-Correction (SC) during inference, we conduct several analytic experiments on COCO-NSS1K and CC-500 under ID and OOD settings. The results are summarized in Tab.~\ref{tab:abla}. 

\vspace{3pt}
\mysubsubsec{Effectiveness of Discriminative Tuning}.
From the compared results in Tab.~\ref{tab:abla} between different variants, we observe that the two tuning objectives, \ie, GM and LG, could consistently promote the alignment performance for T2I according to CLIP and BLIP scores. It verifies the validity of discriminative tuning on ITM and REC tasks. Compared with GM, LG achieves more remarkable improvement over semantic and object detection metrics. It may be attributed to the enhanced grounding ability brought by the prediction of local concepts based on partial descriptions. 
Furthermore, combining the two objectives to conduct multi-task learning may contribute to a slight improvement in BLIP scores under the OOD setting, but other metrics are slightly compromised. This phenomenon indicates that some contradictions may exist during model optimization, reflecting that unifying multiple tasks is still challenging. 

\vspace{3pt}
\mysubsubsec{Effectiveness of Self-Correction}.
In Sec.~\ref{sec:self_corr}, we propose to recycle the discriminative adapter in the inference phase by guiding iterative denoising. Comparing the 3rd and 4th variants in Table~\ref{tab:abla}, we can see that the self-correction scheme could consistently improve the alignment for T2I, attesting to its effectiveness. 

\begin{table}[t]
	\centering
	\setlength{\abovecaptionskip}{0.15cm}
	\caption{Ablation study for the influence of two objectives of discriminative tuning including Global Matching (GM) and Local Ground (LG) in the 2nd stage, and the Self-Correction (SC) during inference on alignment-oriented text-to-image generation. The COCO-NSS1K and CC-500 datasets are used to evaluate in-distribution (ID) and out-of-distribution (OOD) generation. All experiments are based on Stable Diffusion-v2.1.}
    \label{tab:abla}
	{\hspace{-1.3ex}
		\resizebox{0.46\textwidth}{!}
		{
			\setlength\tabcolsep{3.5pt}
			\renewcommand\arraystretch{1.2}
			\begin{tabular}{p{0.8cm}<{\centering}|ccc|ccc|cccc}
				\hline\thickhline
				\multicolumn{1}{c|}{\multirow{2}{*}{Index}} & \multicolumn{1}{c}{ \multirow{2}{*}{GM}} & \multicolumn{1}{c}{ \multirow{2}{*}{LG}} & \multicolumn{1}{c|}{ \multirow{2}{*}{SC}} & \multicolumn{3}{c|}{ COCO-NSS1K (ID)}  & \multicolumn{4}{c}{ CC-500 (OOD)} \\ 
				\cline{5-11}
				&  & & & CLIP & BLIP-M  & BLIP-C & CLIP & BLIP-M  & BLIP-C & GLIP     \\
				\hline\hline
                0       &   &   &   & 34.96  & 73.32  & 40.22  & 39.24  & 85.45  & 43.36  & 52.09  \\
                1       & \Checkmark  &   &   & 35.14  & 74.83  & 40.45  & 39.28  & 86.23  & 43.36  & 49.55  \\
                2       &   & \Checkmark  &   & \textbf{35.94 } & \textbf{79.19 } & 41.11  & \textbf{40.31 } & 90.63  & 44.31  & \textbf{57.03 } \\
                3       &  \Checkmark & \Checkmark  &   & \underline{35.83}  & 78.58  & \underline{41.14} & 40.23  & 90.72  & 44.55  & 53.29  \\
                4       &  \Checkmark & \Checkmark  & \Checkmark  & 35.75  & \underline{79.15}  & \textbf{41.14 } & \underline{40.25}  & \textbf{91.33 } & \textbf{44.69 } & \underline{53.29}  \\
                \hline
			\end{tabular}
		}
	}
\end{table}

\begin{figure}[t]
	\includegraphics[width=0.46\textwidth]{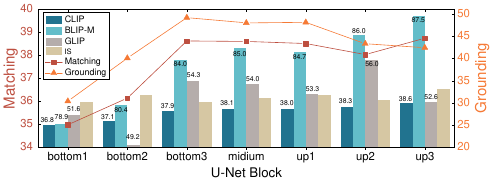}
	\vspace{-2ex}
	\caption{Generative and discriminative results by probing different layers of U-Net in SD-v2.1 and adapting to ITM and REC. We report average CLIP and BLIP-M scores over COCO-NSS1K and CC-500, overall matching performance on MSCOCO-HN, and average grounding performance over all test sets of RefCOCO, RefCOCO+, and RefCOCOg. We conduct model selection based on T2I performance on the validation set of COCO-NSS1K. }
	\label{fig:abla_feat}
	\vspace{-4ex}
\end{figure}

\mysubsubsec{Impact of Probed U-Net Block}.
Due to the hierarchical structure of the U-Net in SD, we could extract multi-level feature maps from its different blocks. Prior work~\cite{xiang2023denoising} has shown that different blocks may have different discriminative powers in image classification. To further investigate the matching and grounding abilities empowered by various blocks and the trade-off between discrimination and generation, we probe consecutive seven blocks of the U-Net shown in Fig.~\ref{fig:pipeline} from left to right and then tune the whole model based on the probed block. The generative and discriminative results are shown in Fig.~\ref{fig:abla_feat}. 
It can be observed that the T2I performance gets continuously improved with the probed block shifting from bottom to up. The reason may be that more LoRA parameters would be introduced and more layers would be tuned during back-propagation. In contrast, the discriminative performance regardless of matching and grounding starts to increase and then deteriorates. It may be attributed to two points: 1) the feature maps from those blocks (\eg, up2 and up3) close to final outputs, \ie, predicted noises, are less semantic; 2) the feature sequences flattened from these feature maps may be too long, making it difficult for the discriminative adapter to probe.

\begin{figure}[h]
\centering
 \subfloat[Tuning Step]{
\label{fig:tuning_step}
\includegraphics[scale=0.15]{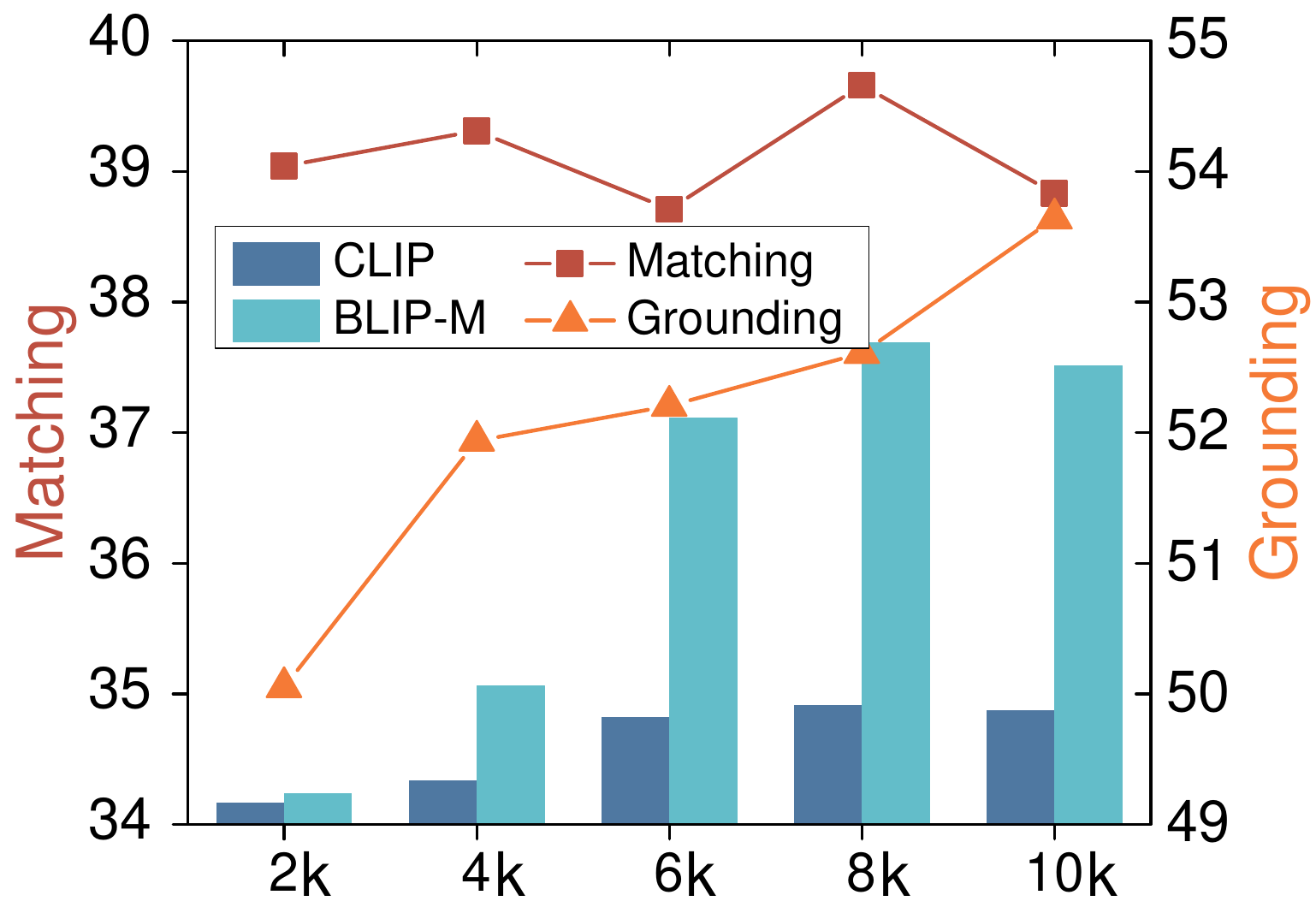}}\hspace{-0.3cm}
\subfloat[Guidance Factor $\eta$]{
\label{fig:self_corr}
\includegraphics[scale=0.15]{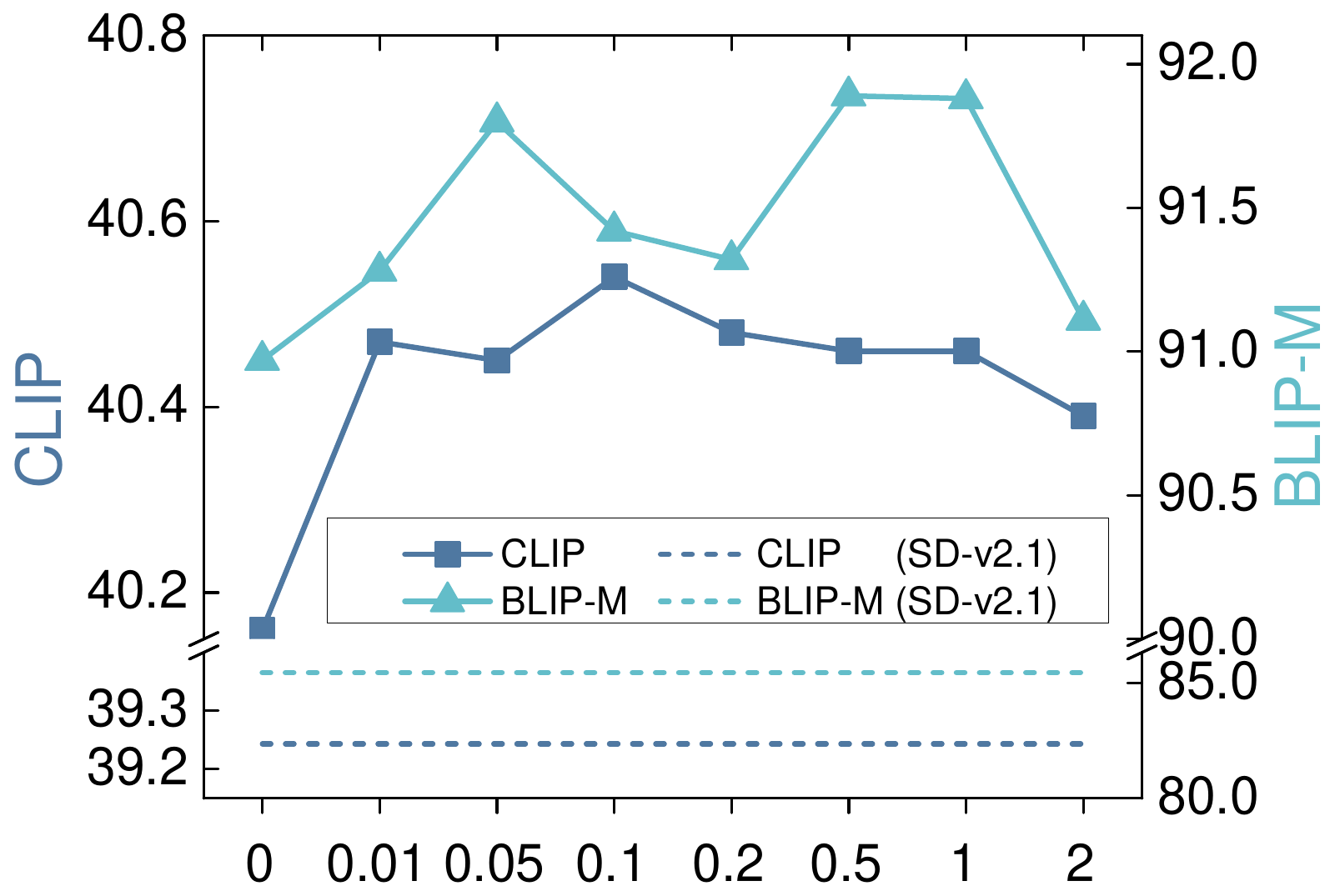}}\vspace{-0.1cm} 
\vspace{-1ex}
\caption{Impact of (a) the variation of generation and discrimination performance with the progress of tuning and (b) the self-correction strength on the performance of T2I on CC-500. }
\label{fig:assign}
\vspace{-2ex}
\end{figure}

\mysubsubsec{Impact of Tuning Step}.
To further delve into the durative impact of discriminative tuning on two aspects of performance, we show the dynamics of the performance with the increment of the tuning step in the 2nd stage in Fig.~\ref{fig:tuning_step}. We can see that the generative performance gets better with tuning and seems to reach the saturation point at the 8k step. In contrast, there is still potential for grounding performance to get higher while the matching performance seems to remain stable in the tuning stage. 

\mysubsubsec{Impact of Self-Correction Factor}.
As shown in Fig.~\ref{fig:self_corr}, we study the influence of the guidance factor $\eta$ in Eqn.~(\ref{eqa:self_corr}) on the alignment performance of T2I. The results demonstrate that the proposed self-correction mechanism could alleviate the text-image misalignment issue with a proper range of guidance factor, \ie, $(0.05, 1)$. 

\begin{figure}[t]
\centering
	\includegraphics[width=0.44\textwidth]{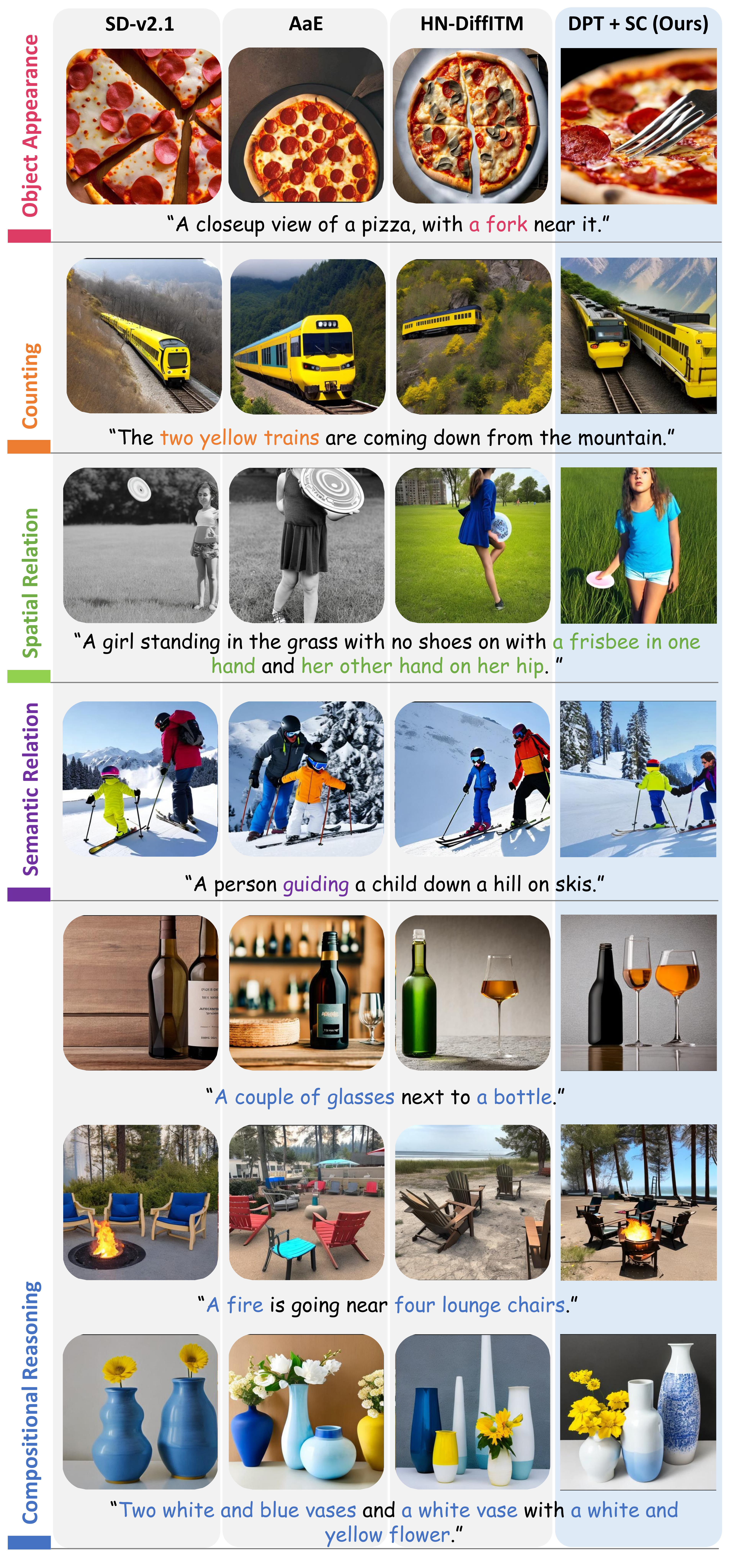}
	\vspace{-2.5ex}
	\caption{Qualitative results on COCO-NSS1K. We compare DPT with SD-v2.1 and two baselines including Attend-and-Excite (AaE)~\cite{chefer2023attend} and HN-DiffusionITM (HN-DiffITM)~\cite{krojer2023diffusion} regarding object appearance, counting, spatial relation, semantic relation, and compositional reasoning. Categories and the corresponding keywords in prompts are highlighted with different colors. }
	\label{fig:case}
	\vspace{-4ex}
\end{figure}

\subsection{Qualitative Results}
To intuitively show the alignment improvement achieved by DPT and SC, we illustrate generated examples with prompts from COCO-NSS1K for object appearance, counting, relation, and compositional reasoning evaluation, as shown in Fig.~\ref{fig:case}. These cases demonstrate the effectiveness of incorporating discriminative probing and tuning into T2I models.

\section{Conclusion and Future Work}
In this work, we tackled the text-image misalignment issue for text-to-image generative models. Toward this end, we retrospected the relations between generative and discriminative modeling and presented a two-stage method named DPT. It introduces a discriminative adapter for probing basic discriminative abilities in the first stage and performs discriminative fine-tuning in the second stage. DPT exhibited effectiveness and generalization across five T2I datasets and four ITM and REC datasets.

In the future, we plan to explore the effect of discriminative probing and tuning to more generative models using more conception and understanding tasks. 
Besides, it is interesting to discuss more complicated relations between discriminative and generative modeling such as trade-offs and mutual promotion across different tasks.  
{
    \small
    \bibliographystyle{ieeenat_fullname}
    \bibliography{main}
}

\clearpage
\appendix
\maketitlesupplementary

\begin{table*}[t]
	\centering
	\setlength{\abovecaptionskip}{0.15cm}
	\caption{Ablation study for the influence of two objectives of discriminative probing and tuning (DPT) and denoising (MSE) on text-to-image generation. The COCO-NSS1K and CC-500 datasets are used to evaluate in-distribution (ID) and out-of-distribution (OOD) generation. Alignment-oriented evaluation metrics include CLIP score~$\uparrow$, BLIP-ITM score~$\uparrow$, and GLIP score~$\uparrow$, while quality-oriented evaluation metrics include IS~$\uparrow$ and FID~$\downarrow$. The best results are highlighted in bold. }
    \label{tab:abla_mse}
	{\hspace{-1.3ex}
		\resizebox{0.95\textwidth}{!}
		{
			\setlength\tabcolsep{10pt}
			\renewcommand\arraystretch{1}
			\begin{tabular}{p{0.8cm}<{\centering}|cc|cccc|cccc}
				\hline\thickhline
				\multicolumn{1}{c|}{\multirow{2}{*}{Version}} & \multicolumn{1}{c}{ \multirow{2}{*}{MSE}} & \multicolumn{1}{c|}{ \multirow{2}{*}{DPT}} & \multicolumn{4}{c|}{ COCO-NSS1K (ID)}  & \multicolumn{4}{c}{ CC-500 (OOD)} \\ 
				\cline{4-11}
				&  & & CLIP & BLIP-ITM  & IS & FID & CLIP & BLIP-ITM  & GLIP & IS     \\
				\hline\hline
                \multicolumn{1}{c|}{\multirow{4}{*}{SD-v1.4}}     &   &    & 33.27  & 67.96  & 31.32  & 54.77  & 34.82  & 70.95  & 31.17  & 14.28  \\
                   & \Checkmark  &    & 32.95  & 70.48  & \textbf{32.03 } & \textbf{52.63 } & 34.00  & 71.27  & 34.08  & \textbf{15.02 } \\
                  &   & \Checkmark   & \textbf{33.85 } & 71.84  & 31.65  & 54.96  & \textbf{35.97 } & 76.74  & 37.07  & 13.56  \\
                 & \Checkmark  & \Checkmark   & 33.83  & \textbf{73.28 } & 30.59  & 55.02  & 35.83  & \textbf{77.20 } & \textbf{37.89 } & 13.67  \\
                \hline
                \multicolumn{1}{c|}{\multirow{4}{*}{SD-v2.1}}     &   &    & 34.96  & 73.32  & 30.40  & 55.35  & 39.24  & 85.45  & 52.09  & 11.53  \\
                   & \Checkmark  &    & 34.20  & 75.90  & 30.10  & \textbf{51.85 } & 38.49  & 84.65  & 49.48  & \textbf{13.38 } \\
                  &   & \Checkmark   & \textbf{35.83 } & 78.58  & \textbf{30.83 } & 55.55  & \textbf{40.23 } & \textbf{90.72 } & \textbf{53.29 } & 11.59  \\
                 & \Checkmark  & \Checkmark   & 35.64  & \textbf{79.23 } & 30.18  & 53.36  & 39.87  & 90.41  & 52.09  & 12.09  \\

                \hline
			\end{tabular}
		}
	}
\end{table*}

\begin{table*}[t]
	\centering
	\setlength{\abovecaptionskip}{0.15cm}
	\caption{Generation performance with different probed blocks of U-Net and the sizes of feature maps (Feat. Size). All the experiments are based on SD-v2.1. We combine multiple feature maps using additive fusion, and perform interpolation if the feature sizes are different.  Alignment-oriented evaluation metrics include CLIP score~$\uparrow$, BLIP-ITM (BLIP-M) score~$\uparrow$, BLIP-ITC (BLIP-C) score~$\uparrow$, and GLIP score~$\uparrow$, while quality-oriented evaluation metrics include IS~$\uparrow$ and FID~$\downarrow$. The best results are highlighted in bold. }
	\label{tab:ablation_blocks}
	{
             \setlength{\tabcolsep}{1mm}{
		\resizebox{0.97\textwidth}{!}
		{
			\setlength\tabcolsep{5pt}
			\renewcommand\arraystretch{1.05}
			\begin{tabular}{lc|ccccc|ccccc}
				\hline\thickhline
				\multicolumn{1}{l}{\multirow{2}{*}{Block}} & \multicolumn{1}{c|}{\multirow{2}{*}{Feat. Size}} & \multicolumn{5}{c|}{ COCO-NSS1K (ID)}  & \multicolumn{5}{c}{ CC-500 (OOD) }  \\ 
				\cline{3-12}
				& & CLIP & BLIP-M  & BLIP-C & \color{gray}{IS} & \color{gray}{FID} & CLIP & BLIP-M & BLIP-C  & GLIP & \color{gray}{IS} \\
				\hline\hline
				-  & - & 34.96  & 73.32  & 40.22  & 30.40  & 55.35  & 39.24  & 85.45  & 43.36  & 52.09  & 11.53  \\
                bottom1  & 32$\times$32 & 34.76  & 73.15  & 40.12  & 29.45  & 55.51  & 38.81  & 84.68  & 43.05  & 51.57  & 11.80  \\
                bottom2  & 16$\times$16 & 35.03  & 73.63  & 40.24  & 30.38  & 55.38  & 39.15  & 87.08  & 43.52  & 49.18  & 12.04  \\
                bottom3  & 8$\times$8 & 35.69  & 77.57  & 40.85  & 30.06  & 55.52  & 40.19  & 90.33  & 44.34  & 54.33  & 11.17  \\
                middle  & 8$\times$8 & 35.90  & 79.25  & 41.11  & 30.53  & 55.12  & 40.28  & 90.66  & 44.40  & 54.04  & 11.31  \\
                up1  & 8$\times$8 & 35.83  & 78.58  & 41.14  & 30.83  & 55.55  & 40.23  & 90.72  & 44.55  & 53.29  & 11.59  \\
                up2  & 16$\times$16 & 35.85  & 79.19  & 41.10  & 30.25  & 54.92  & 40.67  & 92.72  & 44.85  & \textbf{55.98}  & 11.34  \\
                up3  & 32$\times$32 & \textbf{35.91}  & \textbf{80.39}  & \textbf{41.24}  & 31.47  & 57.12  & \textbf{41.21}  & \textbf{94.52}  & \textbf{45.46}  & 52.62  & 11.89  \\
                \hdashline
                middle + bottom3 + up1  & 8$\times$8 & 35.65  & 78.26  & 41.13  & 30.39  & 55.47  & 39.99  & 90.42  & 44.12  & 52.47  & 11.76  \\
                bottom2 + up2  & 16$\times$16 & \textbf{35.91}  & 79.43  & 41.16  & 30.41  & 54.86  & 40.64  & 91.98  & 44.71  & 53.21  & 11.77  \\
                bottom1 + up1  & 32$\times$32 & 35.77  & 79.31  & 41.15  & 30.09  & 56.63  & 40.75  & 93.72  & 45.02  & 51.57  & 12.69  \\
                all  & 8$\times$8 &   35.84  & 79.19  & 41.10  & 30.47  & 55.64  & 40.61  & 91.22  & 44.77  & 53.14  & 11.42  \\
                all  & 32$\times$32 & 35.48  & 78.85  & 41.40  & 30.57  & 56.53  & 40.18  & 92.74  & 45.29  & 48.95  & 11.88  \\
				\hline
			\end{tabular}
		}
            }
	}
\end{table*}

\section{More Results and Analysis}\label{sec:app_more_results}
\subsection{Ablation Study on DPT}
To further explore the effectiveness of the proposed DPT paradigm, we compare it with the traditional denoising tuning method based on the MSE loss function. Concretely, we delve into all four combinations of these two and evaluate the ID and OOD generation performance on the basis of two versions of SD, \ie, SD-v1.4 and SD-v2.1. The experimental results are reported in Tab.~\ref{tab:abla_mse}. We have the following observations. 1) MSE could improve the alignment and quality under the ID setting, but it may not be helpful and even harmful to the alignment under the OOD setting. 2) DPT consistently enhances alignment performance across both ID and OOD settings, with minimal impact on the original image quality. And 3) by combining MSE and DPT objectives, we may get a good trade-off between alignment and quality, and achieve a better performance on the BLIP-ITM evaluation metric which may focus on more details. 

\subsection{Ablation Study on Probed Blocks}
As shown in Tab.~\ref{tab:ablation_blocks}, we carry out extensive experiments to study the effect of different probed blocks of U-Net on generation performance. The results show that DPT achieves the best performance when probing the up3 block.

\subsection{Comprehensive Evaluation on COCO-NSS1K}
\begin{figure*}[t]
\centering
 \subfloat[CLIP Score]{
\label{fig:app_clip_5}
\includegraphics[scale=0.70]{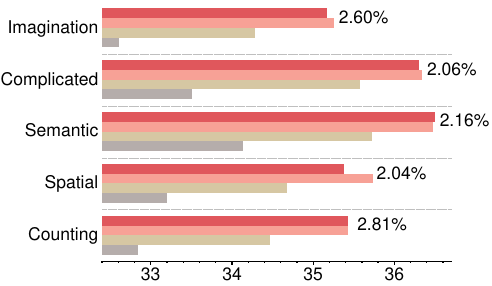}}\hspace{-1cm}
\subfloat[BLIP-ITM Score]{
\label{fig:app_blipm_5}
\includegraphics[scale=0.87]{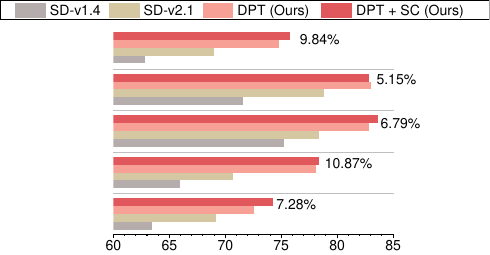}} \hspace{-0.5cm}
\subfloat[BLIP-ITC Score]{
\label{fig:app_blipcc_5}
\includegraphics[scale=0.53]{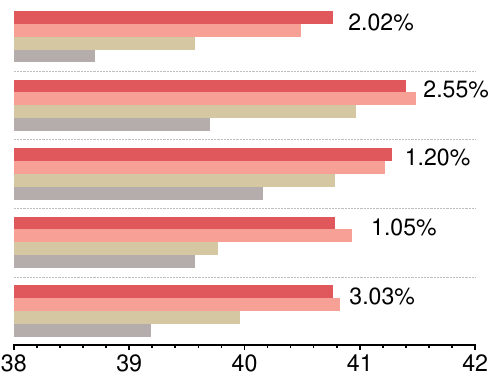}}
\vspace{-1ex}
\caption{Alignment performance improvement of the proposed method compared with SD-v1.4 and SD-v2.1 on five categories of the COCO-NSS1K dataset, including counting, spatial relation, semantic relation, and complicated relation reasoning and imagination abilities evaluation. Results on three evaluation metrics (a) CLIP Score, (b) BLIP-ITM Score, and (c) BLIP-ITC Score are reported. The value on the right of each category denotes the percentage improvement of DPT + SC (Ours) compared to SD-v2.1.}
\label{fig:app_5}
\vspace{-2ex}
\end{figure*}

The COCO-NSS1K dataset~\cite{qu2023layoutllm} was constructed to evaluate five categories of abilities for T2I models, including counting, spatial relation reasoning, semantic relation reasoning, complicated relation reasoning, and abstract imagination. To delve into these categories, we compare the proposed methods, including DPT and DPT + SC, with SD-v1.4 and SD-v2.1, as shown in Fig.~\ref{fig:app_5}. Our method consistently improves the alignment performance in all categories compared with the state-of-the-art SD-v2.1. Besides, the self-correction module could further align the generated images with prompts, especially in the semantic relation category. 

\subsection{Impact of Discriminative Tuning Steps}
\begin{figure}[t]
    \centering
	\includegraphics[width=0.47\textwidth]{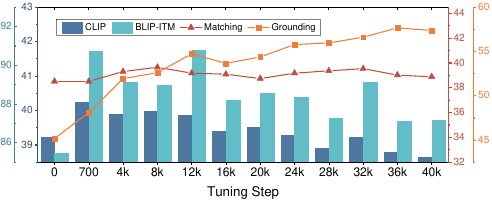}
	\caption{Impact of the discriminative tuning steps in DPT (SD-v2.1) on generation and discrimination performance. CLIP and BLIP-ITM scores on the CC-500 dataset under out-of-distribution setting,  averaging performance over I-to-T and T-to-I matching, and averaging precision@1 over test sets of RefCOCO, RefCOCO+, and RefCOCOg are shown. The model achieves the best generation performance on the validation set at step 700. }
	\label{fig:app_step}
	\vspace{-4ex}
\end{figure}

\begin{figure}[t]
    \centering
	\includegraphics[width=0.47\textwidth]{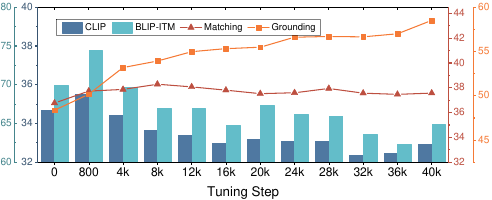}
	\caption{Impact of the discriminative tuning steps in DPT (SD-v1.4) on generation and discrimination performance. CLIP and BLIP-ITM scores on the CC-500 dataset under out-of-distribution setting,  averaging performance over I-to-T and T-to-I matching, and averaging precision@1 over test sets of RefCOCO, RefCOCO+, and RefCOCOg are shown. The model achieves the best generation performance on the validation set at step 700. }
	\label{fig:app_step_sd14}
	\vspace{-4ex}
\end{figure}

\begin{figure}[h]
\centering
 \subfloat[ITM]{
\label{fig:step_itm}
\includegraphics[scale=0.48]{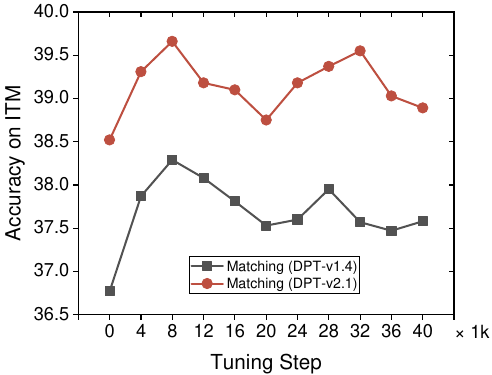}}\hspace{-0.4cm}
\subfloat[REC]{
\label{fig:step_rec}
\includegraphics[scale=0.47]{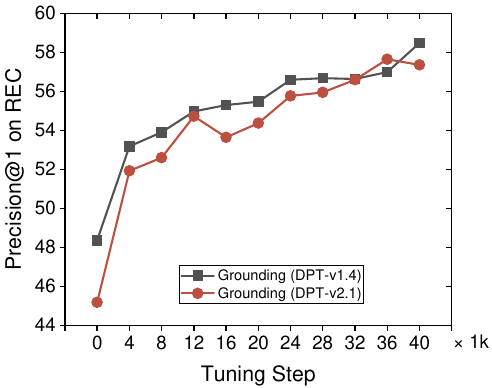}}
\vspace{-1ex}
\caption{Discriminative performance comparison between DPT-v1.4 and DPT-v2.1 on (a) Image-Text Matching (ITM) and (b) Referring Expression Comprehension (REC) as the tuning progresses. }
\label{fig:step_itm_rec}
\vspace{-2ex}
\end{figure}

As shown in Fig.~\ref{fig:app_step}, Fig.~\ref{fig:app_step_sd14}, and Fig.~\ref{fig:step_itm_rec}, we study the generation and discrimination performance based on SD-v2.1 and SD-v1.4, and the comparison between the two
versions, respectively. We discuss the results from the discrimination and generation aspects as follows. 

\mysubsubsec{Discriminative Tuning for Discrimination}
On the one hand, the grounding performance is continuously improved with the tuning step increases, while the matching performance seems to get better at the beginning and then fluctuates within a very small range. Compared with the performance at the end of the discriminative probing phase, \ie, at step 0, the discriminative tuning by introducing the extra parameters of LoRA could further improve both discriminative abilities. On the other hand, the performance on ITM of DPT-v2.1 is obviously better than that of DPT-v1.4. On the contrary, DPT-v1.4 seems to be slightly stronger than DPT-v2.1 in terms of REC, as shown in Fig.~\ref{fig:step_itm_rec}. 

\mysubsubsec{Discriminative Tuning for Generation}
From Fig.~\ref{fig:app_step} and Fig.~\ref{fig:app_step_sd14}, we can see that the performance on discrimination and generation gets better meanwhile at the beginning of tuning, \eg, $[0, 700]$ for DPT-v2.1 and $[0, 800]$ for DPT-v1.4, which demonstrates the effectiveness of discriminative tuning on the enhancement of various abilities. Afterward, we can see that the generation performance declines, perhaps due to over-fitting or some potential discrepancy between generation and discrimination. 

\subsection{Impact of Rank Numbers in LoRA}
The rank number in LoRA determines the number of extra parameters introduced to the discriminative tuning stage compared with the first stage. To explore the influence of the rank numbers on the generation performance, we compare different rank numbers, from 0 to 128, as shown in Tab.~\ref{tab:lora_rank}. The results reflect that DPT achieves the best performance when using 4 rank numbers on most evaluation metrics. More rank numbers do not bring further improvement, which may be attributable to the scale of tuning data. 

\subsection{Impact of Layers of Discriminative Adapter}
The total parameters of DPT also depend on the transformer layers of the discriminative adapter. We conduct experiments by using different numbers of layers. Note that we keep the same number of layers in encoders and decoders for each experiment. The results are reported in Tab.~\ref{tab:layer}. In general, the best generation performance can be achieved when using 4 layers. Besides, the alignment performance is always better than SD-v2.1 (\ie, the experiment with 0 layer), which further verifies the effectiveness of DPT. 

\subsection{Impact of Denosing Objective}
In the raw SD model, only the denoising objective with the MSE form is used to model the data distribution for image synthesis. To further the interplay of DPT and MSE objectives, we perform more experiments by combining them and taking different values of the coefficient of MSE. As shown in Tab.~\ref{tab:coeff_mse}, we observe that the simultaneous use of these two objectives does not cause significant conflict. Instead, there may be a possibility that they can collaborate with each other to achieve a competitive trade-off between text-image alignment and image quality. 

\subsection{Impact of Timesteps on Discriminative Tasks}
As shown in Tab.~\ref{tab:perf_comp_dist}, we explore the influence of different timesteps used in DDPM on the ITM and REC performance. The results illustrate that the proposed model could achieve the best performance when the timestep is set to 250. The performance comparison between 0 and 250 reveals that it is helpful to improve the discriminative abilities by introducing appropriate levels of noise. 

\subsection{Impact of Tunable Modules}
In the second stage, we have two strategies for discriminative tuning: only training LoRA and training DA + LoRA. To make a comparison between the two tuning approaches, we evaluate their generation performance as shown in Tab.~\ref{tab:perf_comp_dist}. From the results, we find training DA + LoRA is better, perhaps due to the more flexibility brought by more parameters from DA and LoRA during the discriminative tuning phase.

\begin{table*}[t]
	\centering
	\setlength{\abovecaptionskip}{0.15cm}
	\caption{Generation performance with different numbers of LoRA ranks on the COCO-NSS1K and CC-500 datasets. Alignment-oriented evaluation metrics include CLIP score~$\uparrow$, BLIP-ITM score~$\uparrow$, BLIP-ITC score~$\uparrow$, and GLIP score~$\uparrow$, while quality-oriented evaluation metrics include IS~$\uparrow$ and FID~$\downarrow$. The best results are highlighted in bold. }
	\label{tab:lora_rank}
	{
             \setlength{\tabcolsep}{1mm}{
		\resizebox{0.97\textwidth}{!}
		{
			\setlength\tabcolsep{9pt}
			\renewcommand\arraystretch{1}
			\begin{tabular}{c|ccccc|ccccc}
				\hline\thickhline
				\multicolumn{1}{c|}{\multirow{2}{*}{Rank}} & \multicolumn{5}{c|}{ COCO-NSS1K (ID)}  & \multicolumn{5}{c}{ CC-500 (OOD) }  \\ 
				\cline{2-11}
				 & CLIP & BLIP-ITM  & BLIP-ITC & \color{gray}{IS} & \color{gray}{FID} & CLIP & BLIP-ITM & BLIP-ITC  & GLIP & \color{gray}{IS} \\
				\hline\hline
				0  & 34.96  & 73.32  & 40.22  & 30.40  & 55.35  & 39.24  & 85.45  & 43.36  & 52.09  & 11.53  \\
                4     & \textbf{35.83 } & \textbf{78.58 } & \textbf{41.14 } & \textbf{30.83 } & 55.55  & \textbf{40.23 } & 90.72  & \textbf{44.55 } & 53.29  & 11.59  \\
                8     & 35.47  & 76.89  & 40.94  & 30.39  & 55.28  & 39.78  & 90.04  & 44.26  & 52.69  & 11.84  \\
                16    & 35.59  & 77.33  & 40.77  & 30.25  & 54.83  & 39.85  & 89.45  & 43.96  & 52.69  & \textbf{11.99 } \\
                32    & 35.55  & 76.68  & 40.80  & 29.98  & \textbf{54.81 } & 39.98  & 89.82  & 44.04  & 52.76  & 11.83  \\
                64    & 35.67  & 77.05  & 40.79  & 30.12  & 55.28  & 40.22  & \textbf{90.99 } & 44.21  & \textbf{53.44 } & 11.73  \\
                128   & 35.66  & 77.11  & 40.84  & 30.68  & 55.34  & 40.05  & 89.99  & 44.12  & 52.39  & 11.92  \\
				\hline
			\end{tabular}
		}
            }
	}
\end{table*}

\begin{table*}[t]
	\centering
	\setlength{\abovecaptionskip}{0.15cm}
	\caption{Generation performance with different numbers of layers of encoders and decoders in the discriminative adapter on the COCO-NSS1K and CC-500 datasets. Alignment-oriented evaluation metrics include CLIP score~$\uparrow$, BLIP-ITM score~$\uparrow$, BLIP-ITC score~$\uparrow$, and GLIP score~$\uparrow$, while quality-oriented evaluation metrics include IS~$\uparrow$ and FID~$\downarrow$. The best results are highlighted in bold. }
	\label{tab:layer}
	{
             \setlength{\tabcolsep}{1mm}{
		\resizebox{0.97\textwidth}{!}
		{
			\setlength\tabcolsep{9pt}
			\renewcommand\arraystretch{1}
			\begin{tabular}{c|ccccc|ccccc}
				\hline\thickhline
				\multicolumn{1}{c|}{\multirow{2}{*}{Layer}} & \multicolumn{5}{c|}{ COCO-NSS1K (ID)}  & \multicolumn{5}{c}{ CC-500 (OOD) }  \\ 
				\cline{2-11}
				 & CLIP & BLIP-ITM  & BLIP-ITC & \color{gray}{IS} & \color{gray}{FID} & CLIP & BLIP-ITM & BLIP-ITC  & GLIP & \color{gray}{IS} \\
				\hline\hline
				0     & 34.96  & 73.32  & 40.22  & 30.40  & 55.35  & 39.24  & 85.45  & 43.36  & 52.09  & 11.53  \\
                1     & 35.83  & 78.58  & \textbf{41.14 } & 30.83  & 55.55  & 40.23  & 90.72  & 44.55  & 53.29  & 11.59  \\
                2     & 35.51  & 77.67  & 40.93  & 30.52  & \textbf{54.75 } & 40.23  & 91.14  & 44.36  & 52.99  & 11.65  \\
                3     & 35.55  & 77.14  & 40.81  & 30.27  & 55.02  & 40.03  & 90.55  & 44.19  & 53.59  & \textbf{12.00 } \\
                4     & \textbf{35.91 } & \textbf{79.41 } & 41.08  & \textbf{30.95 } & 55.06  & \textbf{40.88 } & \textbf{92.92 } & \textbf{44.88 } & 54.26  & 11.53  \\
                5     & 35.80  & 78.76  & 41.02  & 29.53  & 55.18  & 40.10  & 90.89  & 44.22  & \textbf{55.23 } & 11.97  \\
				\hline
			\end{tabular}
		}
            }
	}
\end{table*}

\begin{table*}[t]
	\centering
	\setlength{\abovecaptionskip}{0.15cm}
	\caption{Generation performance with different coefficients (Coeff.) of the denoising MSE loss function at the second stage for the discriminative tuning on the COCO-NSS1K and CC-500 datasets. Alignment-oriented evaluation metrics include CLIP score~$\uparrow$, BLIP-ITM score~$\uparrow$, BLIP-ITC score~$\uparrow$, and GLIP score~$\uparrow$, while quality-oriented evaluation metrics include IS~$\uparrow$ and FID~$\downarrow$. The best results are highlighted in bold. }
	\label{tab:coeff_mse}
	{
        \setlength{\tabcolsep}{1mm}{
		\resizebox{0.97\textwidth}{!}
		{
			\setlength\tabcolsep{8pt}
			\renewcommand\arraystretch{1}
			\begin{tabular}{c|ccccc|ccccc}
				\hline\thickhline
				\multicolumn{1}{c|}{\multirow{2}{*}{Coeff. of MSE}} & \multicolumn{5}{c|}{ COCO-NSS1K (ID)}  & \multicolumn{5}{c}{ CC-500 (OOD) }  \\ 
				\cline{2-11}
				 & CLIP & BLIP-ITM  & BLIP-ITC & \color{gray}{IS} & \color{gray}{FID} & CLIP & BLIP-ITM & BLIP-ITC  & GLIP & \color{gray}{IS} \\
				\hline\hline
				0.00  & \textbf{35.83 } & 78.58  & 41.14  & \textbf{30.83 } & 55.55  & \textbf{40.23 } & \textbf{90.72 } & 44.55  & 53.29  & 11.59  \\
                0.05  & 35.54  & 77.78  & 41.14  & 29.73  & 52.84  & 40.05  & 90.12  & 44.41  & 52.84  & 11.57  \\
                0.10  & 35.64  & 79.08  & 41.34  & 29.74  & 52.79  & 40.11  & 89.85  & 44.37  & 53.36  & 11.97  \\
                0.30  & 35.53  & 78.81  & 41.32  & 29.95  & \textbf{52.73 } & 40.03  & 89.65  & 44.37  & \textbf{54.48 } & \textbf{12.20 } \\
                0.50  & 35.49  & 77.51  & 40.99  & 30.61  & 53.44  & 39.81  & 88.78  & 44.05  & 50.90  & 11.26  \\
                1.00  & 35.64  & \textbf{79.23 } & \textbf{41.49 } & 30.18  & 53.36  & 39.87  & 90.41  & \textbf{44.70 } & 52.09  & 12.09  \\
				\hline
			\end{tabular}
		}
            }
	}
\end{table*}

\begin{table*}[t]
	\centering
	\setlength{\abovecaptionskip}{0.15cm}
	\caption{Discriminative performance comparison with different timesteps for \textit{image-text matching} and \textit{referring expression comprehension} to evaluate global matching and local grounding abilities, respectively. Larger timesteps mean stronger noises are introduced to contaminate input images. Datasets include MSCOCO-HN for ITM, and RefCOCO, RefCOCO+, and RefCOCOg for REC. }
    \label{tab:perf_comp_dist}
	{\hspace{-1.3ex}
        \setlength{\tabcolsep}{0.5mm}{
		\resizebox{\textwidth}{!}
		{
			\setlength\tabcolsep{13pt}
			\renewcommand\arraystretch{1}
			\begin{tabular}{c|ccc|cc:cc:c:c}
				\hline\thickhline
				\multicolumn{1}{c|}{\multirow{2}{*}{Method}} & \multicolumn{3}{c|}{ MSCOCO-HN} & \multicolumn{2}{c:}{ RefCOCO}  & \multicolumn{2}{c:}{ RefCOCO+} & \multicolumn{1}{c:}{ RefCOCOg} & \multicolumn{1}{c}{\multirow{2}{*}{Avg.}}\\ %
				\cline{2-9}
				& I-to-T & T-to-I & Overall & testA  & testB & testA & testB & test &   \\
				\hline\hline
                Random Chance & 25.00  & 25.00  & 25.00  & 13.51  & 19.20  & 13.57  & 19.60  & 19.10  & 17.00  \\
                0     & 44.02  & 34.36  & 39.19  & 60.53  & 49.32  & 54.16  & 42.38  & 52.74  & 51.83  \\
                250   & \textbf{44.31 } & \textbf{35.13 } & \textbf{39.72 } & \textbf{63.80 } & \textbf{52.03 } & \textbf{56.72 } & \textbf{44.04 } & \textbf{55.03 } & \textbf{54.32 } \\
                500   & 42.07  & 34.97  & 38.52  & 57.84  & 46.73  & 50.12  & 38.41  & 47.45  & 48.11  \\
                750   & 38.09  & 32.13  & 35.11  & 35.90  & 27.81  & 28.50  & 21.50  & 28.00  & 28.34  \\
                999   & 34.48  & 28.56  & 31.52  & 19.50  & 15.17  & 13.31  & 9.10  & 9.10  & 13.24  \\
                \hline
			\end{tabular}
		}}
	}
         \vspace{-0.2cm}
\end{table*}

\begin{table}[t]
	\centering
	\setlength{\abovecaptionskip}{0.15cm}
	\caption{Generation performance comparison with different tuning strategies at the second stage. ``LoRA'' refers to only training LoRA and freezing other modules including U-Net and Discriminative Adapter, while ``DA + LoRA'' means training Discriminative Adapter and LoRA with only U-Net frozen. }
    \label{tab:perf_comp_dist}
	{\hspace{-1.3ex}
        \setlength{\tabcolsep}{0.5mm}{
		\resizebox{0.45\textwidth}{!}
		{
			\setlength\tabcolsep{7pt}
			\renewcommand\arraystretch{1}
			\begin{tabular}{c|cc|cc}
				\hline\thickhline
				\multicolumn{1}{c|}{\multirow{2}{*}{Trainable}} & \multicolumn{2}{c|}{COCO-NSS1K} & \multicolumn{2}{c}{CC-500}   \\ 
				\cline{2-5}
				& CLIP & BLIP-ITM & CLIP & BLIP-ITM      \\
				\hline\hline
                LoRA  & 35.48  & 76.46  & 39.75  & 89.23  \\
                DA + LoRA & \textbf{35.83}  & \textbf{78.58}  & \textbf{40.23}  & \textbf{90.72}  \\
                \hline
			\end{tabular}
		}}
	}
         \vspace{-0.2cm}
\end{table}

\section{More Experimental Settings}\label{sec:app_exp_settings}

\subsection{Benchmark Datasets}\label{sec:app_benchmarks}
\mysubsubsec{Data used for Training}.
We evaluate the basic global matching and local grounding abilities of T2I models based on two discriminative tasks, \ie, Image-Text Matching and Referring Expression Comprehension, respectively. For discriminative probing and tuning, we reorganize public benchmarks including MSCOCO~\cite{lin2014microsoft} for ITM, and RefCOCO~\cite{yu2016modeling}, RefCOCO+~\cite{yu2016modeling}, and RefCOCOg~\cite{yu2016modeling} for REC. 
Specifically, we use the COCO2014 version of MS-COCO, composed of 82,783 images and 414,113 captions in the training set. There are about 5 caption annotations for each image. As for REC, following MDETR~\cite{kamath2021mdetr}, we combine the three datasets, \ie, RefCOCO, RefCOCO+, and RefCOCOg, into one, called RefCOCOall in the following. Its training set includes 28,158 images and 321,327 expressions. 

To probe global matching and local grounding abilities at the same time, we combine all the above datasets into one. Concretely, we observe that the MSCOCO dataset includes all the raw images in the above three REC datasets. Therefore, we adopt the following data sampling~\cite{nie2022search} strategy during training: 1) randomly sample an expression $y'$ from RefCOCOall and get the corresponding image $x$, 2) randomly sample a caption $y$ from all positive captions of $x$, 3) randomly sample a hard negative caption $y^{neg}$ from top-20 hard negative captions\footnote{We use OpenCLIP (ViT-H-14) to calculate image-text similarities and retrieve the top-k hard negative captions (k=20) or hard negative images (k=4).} of $x$ from the training set of MS-COCO, 4) randomly sample a hard negative image $x^{neg}$ from top-4 hard negative images of $y$ from the training set of MS-COCO, and 5) randomly sample a negative image $x^{rand}$ and a negative caption $y^{rand}$ from all other uncorrelated images and captions, respectively,  from the training set of MS-COCO. Finally, we have a septuple $(x, y', y, y^{neg}, x^{neg}, y^{rand}, x^{rand})$ in each data instance. 

\mysubsubsec{Benchmarks for Evaluation of Generation}. 
Originating from MSCOCO~\cite{lin2014microsoft}, \textbf{COCO-NSS1K}~\cite{qu2023layoutllm} is specially reorganized to assess counting and relation understanding of generative models in complex scenes, including 943 natural prompts and relevant ground-truth images. 
\textbf{CC-500}~\cite{feng2022training} is built to evaluate compositional generation by template-based 446 prompts that conjunct two concepts. 
\textbf{ABC-6K}~\cite{feng2022training} is composed of 6,434 prompts, including 3,217 natural prompts from MSCOCO in which at least two color words describe different objects, and the other 3,217 prompts obtained by switching the positions of two color words. 
\textbf{TIFA}~\cite{hu2023tifa} is a recent VQA-based benchmark built to evaluate T2I alignment across 12 categories, consisting of 4,081 prompts from MSCOCO~\cite{lin2014microsoft}, DrawBench~\cite{saharia2022photorealistic}, PartiPrompt~\cite{yu2022scaling}, and PaintSill~\cite{cho2023dall}. The VQA accuracy based on MLLMs~\cite{li2022mplug} is employed to assess text-image alignment. 
As a contemporary work, \textbf{T2I-CompBench}~\cite{huang2023t2i} is constructed to offer a comprehensive benchmark for compositional T2I from 6 categories, including color binding, shape binding, texture binding, spatial relationships, non-spatial relationships, and complex compositions. We use the test set with 1,800 prompts for evaluation.

\mysubsubsec{Benchmarks for Evaluation of Discrimination}. 
To evaluate the discriminative global matching~\cite{qu2020context, wen2023target, wen2021comprehensive, sun2022counterfactual} ability of different T2I models, we reorganize the test sets of MS-COCO for efficient evaluation. Specifically, as for the Image-to-Text (I-to-T) matching direction, we first retrieve the top-20 hard negative captions and then randomly sample 3 captions. In this way, we have one positive captions and three negative captions for each image. Similarly, we randomly sample 3 negative images from the top-4 retrieval results for the T-to-I matching direction. We name this hard-negative-based test set as \textit{HN-MSCOCO}. As for local grounding, we evaluate generative models following existing methods~\cite{kamath2021mdetr, subramanian2022reclip, liu2023dq} for REC.

\subsection{Baselines}\label{sec:app_baselines}
To verify the effectiveness of our method on T2I, we carry out experiments based on SD-v1.4 and SD-v2.1, and compare DPT with SD~\cite{rombach2022high} and recent alignment-oriented T2I baselines
including LayoutLLM-T2I~\cite{qu2023layoutllm}, StructureDiffusion~\cite{feng2022training}, Attend-and-Exite~\cite{chefer2023attend}, DiffusinoITM~\cite{krojer2023diffusion}, VPGen~\cite{cho2023visual} and LayoutGPT~\cite{feng2023layoutgpt}. 

\subsection{Implementation Details}\label{sec:app_imple_details}
\subsubsection{Settings for Discriminative Probing and Tuning} 
In the first stage, we perform discriminative tuning by pre-training for 60k steps with a batch size of 220 and a learning rate of 1e-4. After that, we perform discriminative tuning in the second stage for 6k steps with a batch size of 8 and a learning rate of 1e-4. We use the AdamW~\cite{loshchilov2017decoupled} optimizer with betas as $(0.9, 0.999)$ and weight decay as 0.01. We perform the gradient clip with 0.1 as the maximum norm. 
We select the model checkpoint by using a validation set for alignment-oriented generation. Specifically, we collect this validation set from MS-COCO, following COCO-NSS1K~\cite{qu2023layoutllm}. Based on this set, we generate images conditioned on prompts and then compute the CLIP score between them. The model with the highest CLIP score in this validation set is chosen for testing. 
Besides, we accumulate gradients in each 8 steps in this stage. Using a single A100 (40G), the discriminative probing requires about 3 days for the first stage, and the discriminative tuning requires about 1 day for the second stage. 

\subsubsection{Settings for Discriminative Adapter} 
We use the 1-layer Transformer encoder and the 1-layer Transformer decoder to implement the discriminative adapter introduced in Sec.~\ref{sec:dis_prob}. The feature map with the shape $1280\times8\times8$ in the medium layer of U-Net is flattened and fed into the encoder. The hidden dimensions of attention layers and feedforward layers are set to 256 and 2048, respectively. The number of attention heads is 8. We use 110 learnable queries in total, 10 for global matching, and the other 100 for local grounding, \ie, $N=110$ and $M=10$. 
The temperature factor $\tau$ for contrastive learning in Eqn.~(\ref{eqn:loss_t2i}) and Eqn.~(\ref{eqn:loss_i2t}) is learnable and initialized to 0.07. 
To make a balance between different grounding objectives in Eqn.~(\ref{eqn:grounding_loss}), we set $\lambda_0$, $\lambda_1$, $\lambda_2$, and $\lambda_3$ as 1, 5, 2, and 1, respectively. The same setting is also used for maximum matching in Eqn.~(\ref{eqn:max_matching}). During inference, we use 0.5 as the default guidance strength of self-correction, \ie, $\eta = 0.5$. 

\subsubsection{Implementation of Baselines}
We run the codes of SD-v1.4\footnote{\url{https://huggingface.co/CompVis/stable-diffusion-v1-4}.} and SD-v2.1\footnote{\url{https://huggingface.co/stabilityai/stable-diffusion-2-1}.} in the huggingface open-source community. Besides, we run the code of LayoutLLM-T2I\footnote{\url{https://github.com/LayoutLLM-T2I/LayoutLLM-T2I}.} and load the open checkpoint to evaluate its performance on COCO-NSS1K. Because of the training-free property, StructureDiffusion\footnote{The code of StructureDiffusion can be found at \url{https://github.com/weixi-feng/Structured-Diffusion-Guidance}, but it is only implemented based on SD-v1.4.} and Attend-and-Excite\footnote{\url{https://github.com/yuval-alaluf/Attend-and-Excite}.} can be directly executed for evaluation. As for DiffusionITM~\cite{krojer2023diffusion}, we implement it based on the open-sourced code\footnote{\url{https://github.com/McGill-NLP/diffusion-itm}.} and rename it as HN-DiffusionITM considering we perform contrastive learning based on the hard negative samples. For a fair comparison, we re-train HN-DiffusionITM using our training data and adopt the DDIM sampler with 50 timesteps to synthesize images. 

To compare the global matching abilities of different generative models, we implement Diffusion Classifier~\cite{li2023your} and DiffusionITM~\cite{krojer2023diffusion}. They both depend on ELBO to compute text-image similarities. Compared with Diffusion Classifier, DiffusionITM uses a normalization method to rectify the text-to-image directional matching, dealing with the modality asymmetry issue to some extent. We call this version as HN-DiffusionITM.

Regarding the local grounding ability, we first evaluate CLIP and OpenCLIP by means of the cropping strategy~\cite{subramanian2022reclip}. In detail, we first crop the raw images into multiple blocks and resize them to the same size according to the proposals, and then perform expression-to-image matching. In addition, we also adopt the cropping and expression-to-image matching strategy to evaluate the grounding performance of Diffusion Classifier, DiffusionITM, and HN-DiffusionITM, since these models can not be directly repurposed to do the REC task. We also propose a local denoising method based on Diffusion Classifier. Specifically, we calculate the ELBO for each local proposal region instead of the whole image, and then take the proposal with the maximum ELBO values as the prediction. 

\section{More Examples}\label{sec:app_more_example}
\begin{figure}[htbp]
    \centering
	\includegraphics[width=0.48\textwidth]{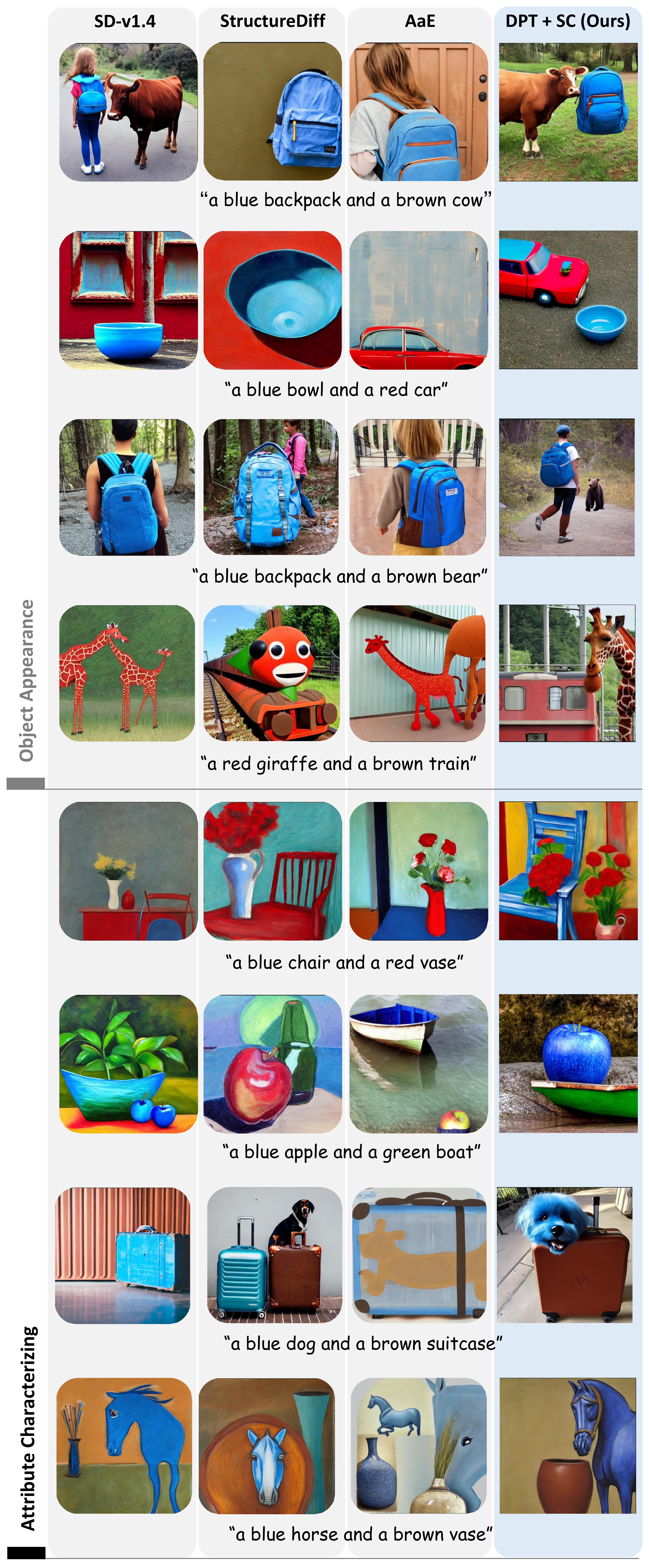}
	\caption{Qualitative results on CC-500 We compare the proposed method with SD-v1.4 and two baselines including StructureDiff~\cite{feng2022training} and Attend-and-Excite (AaE)~\cite{chefer2023attend} regarding object appearance and attribute characterizing. }
	\label{fig:case_cc500}
\end{figure}

\begin{figure}[t]
    \centering
	\includegraphics[width=0.485\textwidth]{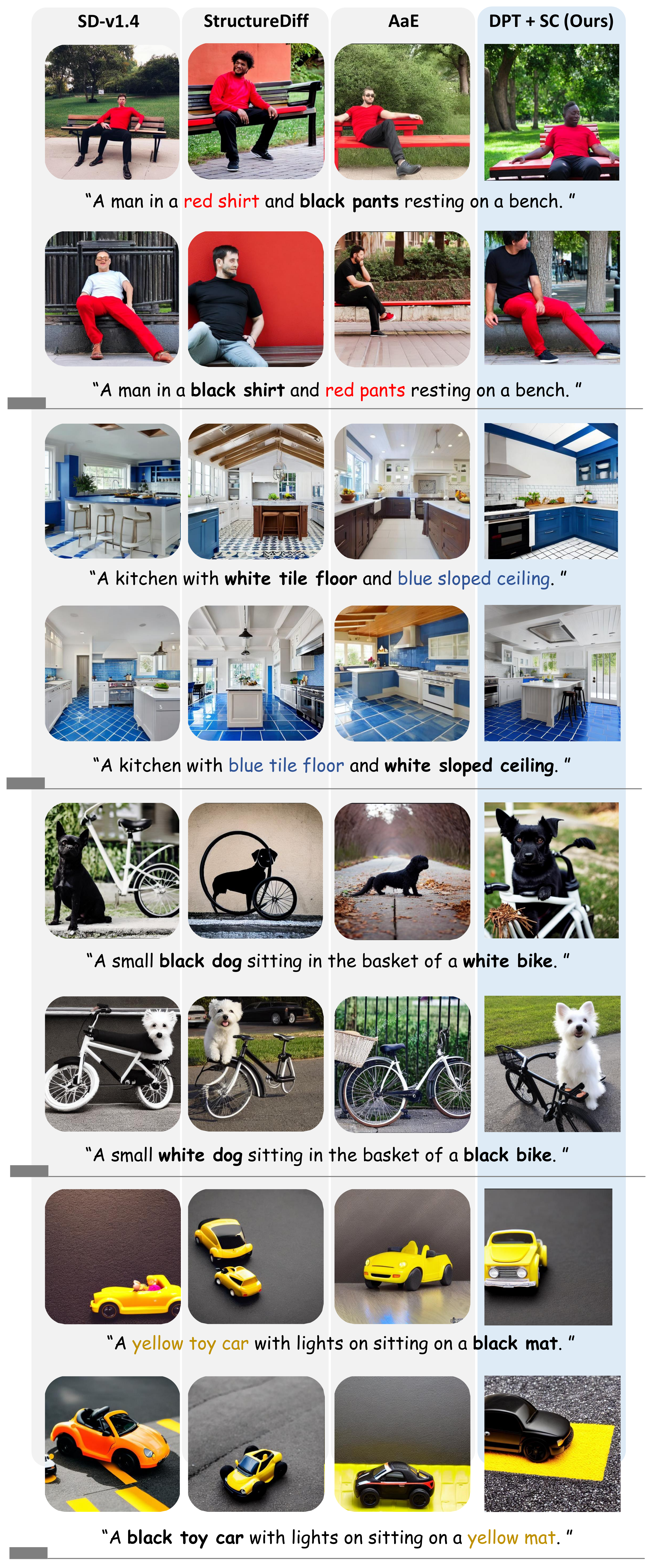}
	\caption{Qualitative results on ABC-6K. We compare the proposed method with SD-v1.4 and two baselines including StructureDiff~\cite{feng2022training} and Attend-and-Excite (AaE)~\cite{chefer2023attend} regarding color attribute characterizing. }
	\label{fig:case_abc-6k}
\end{figure}

\begin{figure*}[t]
    \centering
    \vspace{-3ex}
	\includegraphics[width=0.93\textwidth]{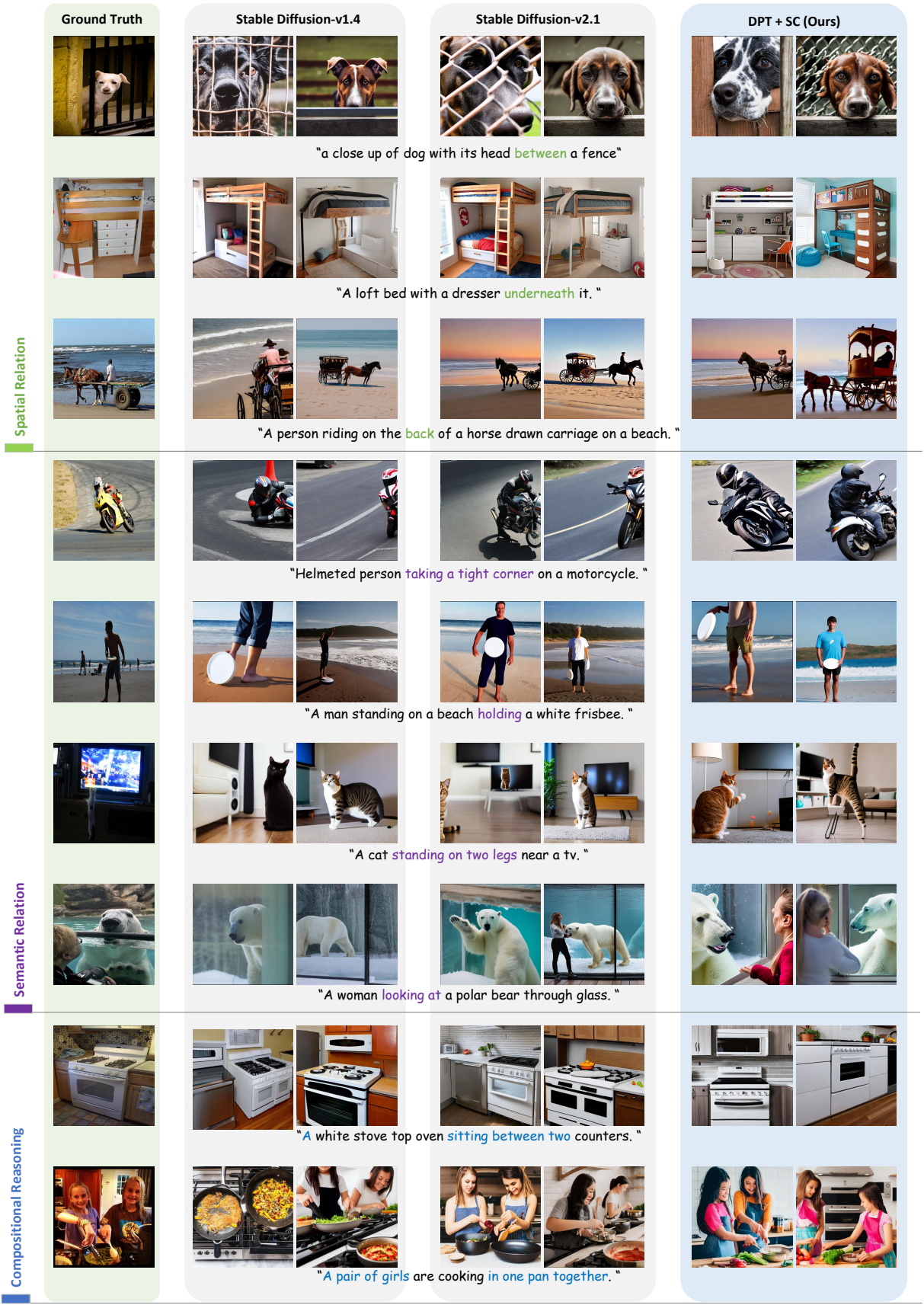}
	\vspace{-2ex}
	\caption{More generated examples on COCO-NSS1K regarding spatial, semantic, and compositional reasoning.}
 \label{fig:nss_sd14_sd21_ours_relation}
\end{figure*}

\begin{figure*}[t]
    \centering
    \vspace{-3ex}
	\includegraphics[width=0.90\textwidth]{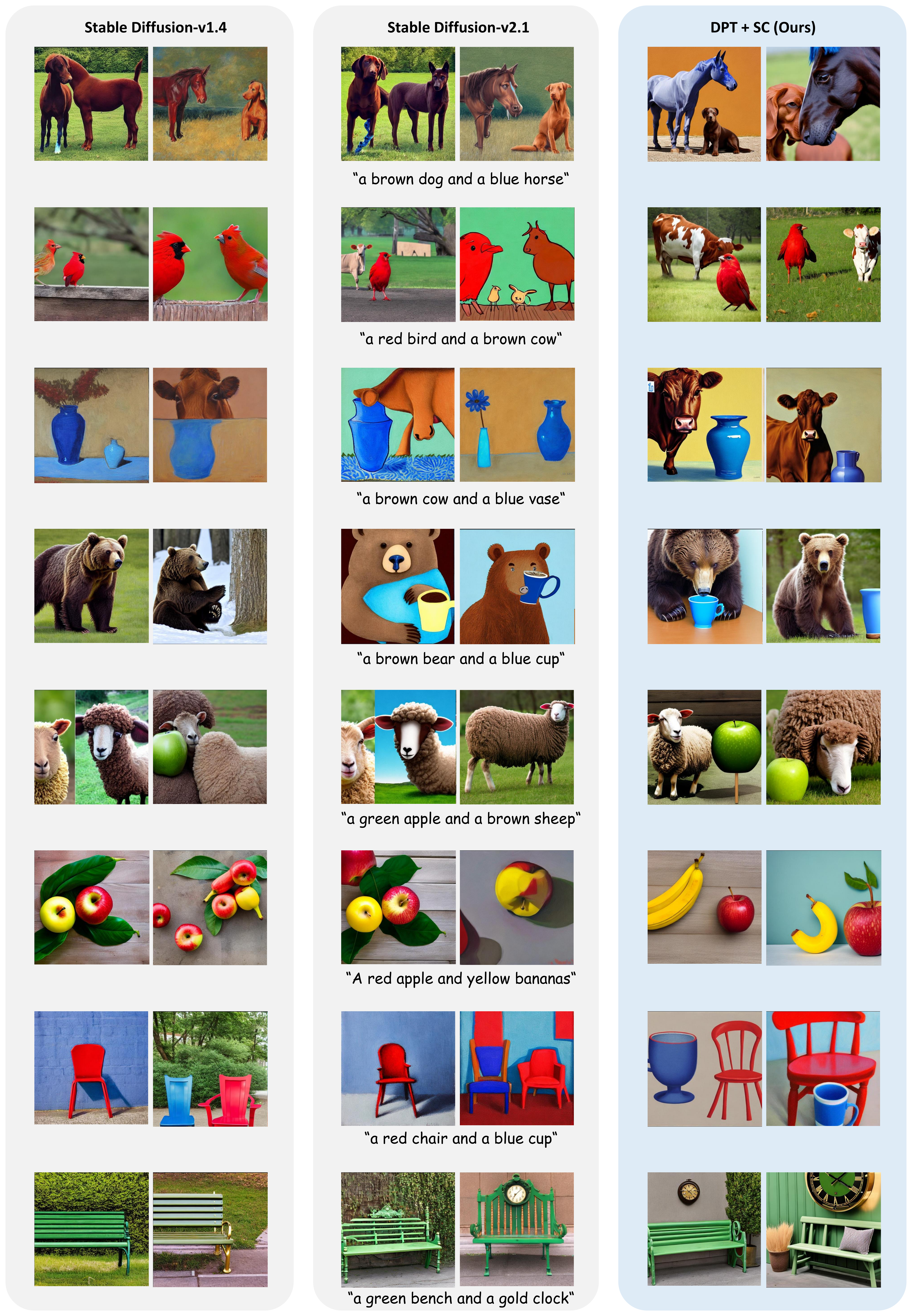}
	\vspace{-2ex}
	\caption{More generated examples on the CC-500 dataset.}
 \label{fig:cc500_sd14_sd21_ours}
\end{figure*}

\begin{figure*}[t]
    \centering
    \vspace{-3ex}
	\includegraphics[width=0.90\textwidth]{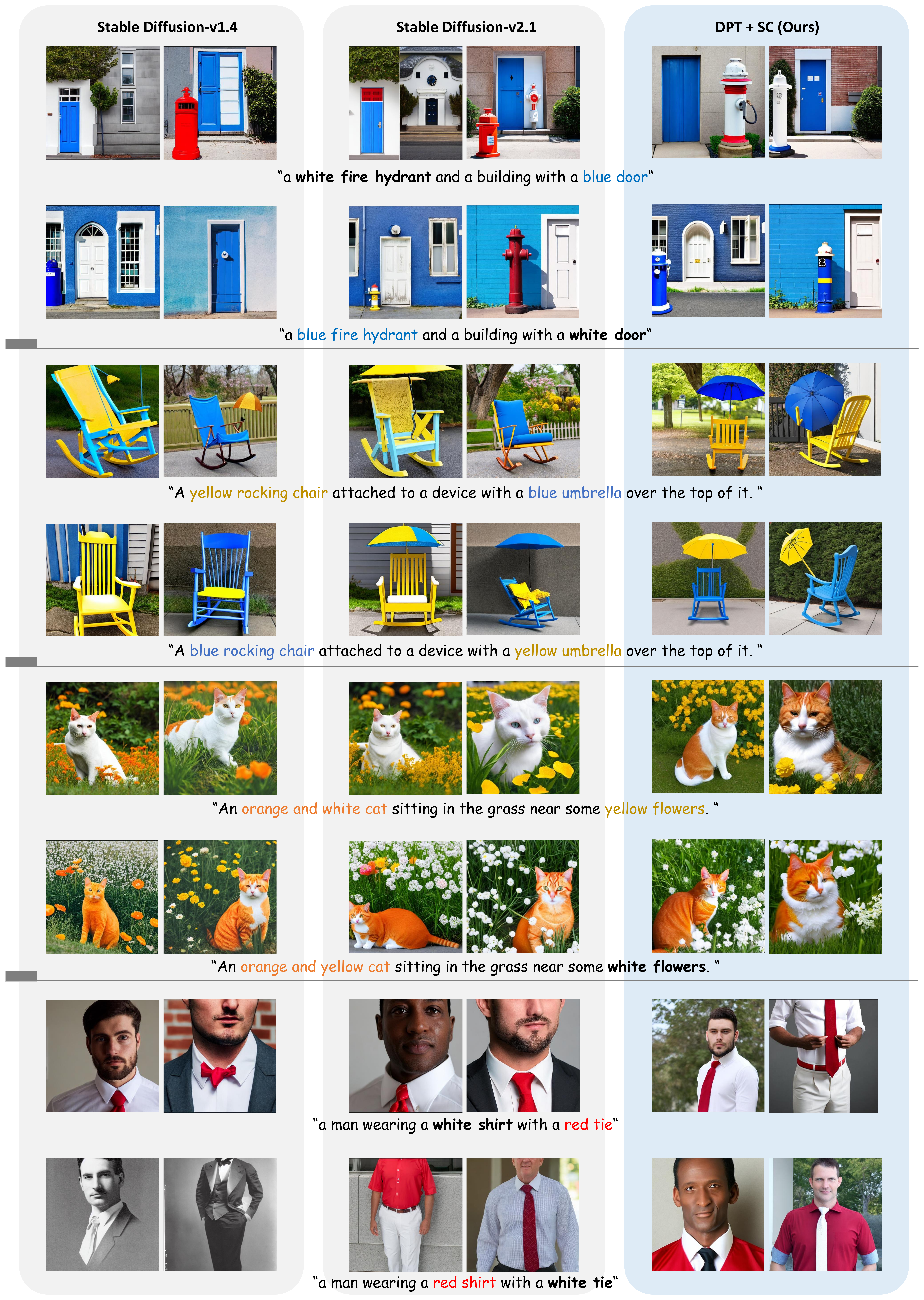}
	\vspace{-2ex}
	\caption{More generated examples on the ABC-6K dataset.}
 \label{fig:abc6k_sd14_sd21_ours}
\end{figure*}

To intuitively compare the proposed method with recent baselines, we list some generated images on the CC-500 and ABC-6K datasets, as shown in Fig.~\ref{fig:case_cc500} and Fig.~\ref{fig:case_abc-6k}, respectively. To make a fair comparison, we implement all the methods based on SD-v1.4, considering that StructureDiffusion only supports SD-v1.4. Besides, we also show more diverse examples including SD-v1.4, SD-v2.1, and ours, on the COCO-NSS1K, CC-500, and ABC-6K datasets, in Fig.~\ref{fig:nss_sd14_sd21_ours_appa_count} and Fig.~\ref{fig:nss_sd14_sd21_ours_relation}, Fig.~\ref{fig:cc500_sd14_sd21_ours}, and Fig.~\ref{fig:abc6k_sd14_sd21_ours}, respectively. 

\clearpage

\begin{figure*}[t]
    \centering
    \vspace{-3ex}
	\includegraphics[width=0.93\textwidth]{fig/nss_sd14_sd21_ours_appa_count.pdf}
	\vspace{-2ex}
	\caption{More generated examples on COCO-NSS1K regarding object appearance and counting.}
    \label{fig:nss_sd14_sd21_ours_appa_count}
\end{figure*}

\end{document}